\documentclass{article}

\usepackage[main, final]{neurips_2026}

\usepackage[utf8]{inputenc} 
\usepackage[T1]{fontenc}    
\usepackage{hyperref}       
\usepackage{url}            
\usepackage{booktabs}       
\usepackage{amsfonts, amsmath, amsthm, amssymb, comment}       
\usepackage{nicefrac}       
\usepackage{microtype}      
\usepackage{xcolor}         
\usepackage{natbib}

\usepackage{algorithm2e}

\usepackage{booktabs}
\usepackage{caption}

\usepackage{algorithm2e}
\SetAlgoLined
\SetAlgoVlined
\LinesNumbered

\usepackage{tcolorbox}
\tcbuselibrary{breakable}
\usepackage[capitalize,noabbrev]{cleveref}

\usepackage{tikz}
\usetikzlibrary{
  arrows.meta,
  positioning,
  shapes.geometric,
  shapes.misc,
  calc,
  decorations.pathreplacing
}
\tikzset{line/.style={-latex, thick}}

\newtheorem{theorem}{Theorem}[section]
\newtheorem{proposition}[theorem]{Proposition}
\newtheorem{lemma}[theorem]{Lemma}
\newtheorem{corollary}[theorem]{Corollary}

\newtheorem{assumption}[theorem]{Assumption}

\title{DUST: Directional Uncertainty-aware and Scale-invariant Transfer Learning}

\author{%
  Abhisek Chakraborty\thanks{There was no external or internal funding for this work. } \\
  Department of Biostatistics\\
  Johns Hopkins Bloomberg School of Public Health\\
  Baltimore, MD 21205, United States\\
  \texttt{zovialpapai@gmail.com} \\
}

\begin{document}

\maketitle

\begin{abstract}
Transfer learning seeks to improve  efficiency in a target regression task by borrowing information from related external source dataset(s). Existing approaches achieve this by enforcing Euclidean proximity between the target and source regression parameters or, more recently, by encouraging  their directional alignment through angle-based penalties. While such angle-based penalties mitigate the sensitivity of the transfer learning methods to the scale differences between source and target regression coefficients, they still rely on a point estimate of the source regression parameter and thus provide no direct mechanism to incorporate uncertainty in the source information. To that end, we propose a probabilistic framework for transfer learning that operates at the level of a \emph{scale-invariant directional distribution} of the source regression parameter estimate, rather than its point estimate, to enable robust, scale-invariant and uncertainty aware transfer under heterogeneous source information. Specifically, we project the sampling distribution of the source regression estimate onto the unit sphere, thereby extracting its scale-free directional distribution. Transfer from source to target is then induced through a novel \emph{angular cone prior} which shrinks the target regression parameter towards directions in which the source distribution assigns high probability, while allowing its magnitude to be learned exclusively from the target dataset.  The proposed construction admits exact finite-sample characterizations of posterior angular concentration and posterior computation remains tractable via careful low-dimensional augmentation in the parameter space. Simulation studies and application on a benchmark disease status prediction task based on the National Health and Nutrition Examination Survey (NHANES) datasets demonstrate that the proposed method yields marked gains over existing approaches in settings with substantial uncertainty in the source regression estimate and source--target scale mismatch. 
\end{abstract}

\section{Introduction}

Transfer learning \citep{pan2010survey, ben2007analysis, ben2010theory, li2018transfer, cai2022transfer,  suder2025bayesian} has emerged as a powerful framework for enhancing  efficiency in statistical inference when the target dataset is limited in size but external information is available through related source dataset(s). In regression settings, such external information often takes the form of a fitted model or regression parameter estimate obtained from the source dataset. The key challenge in transfer learning is then to borrow strength from the source dataset while remaining robust to heterogeneity between the source and target datasets. A substantial body of work tackles the problem of  transfer learning by utilizing penalties that enforce Euclidean proximity between the target and source parameters via distance-based regularization \citep{evgeniou2004regularized, suder2025bayesian}, or Bayesian hierarchical priors to borrow strength by shrinking the target coefficient vector towards a source estimate \citep{datta2021regularized,  shwartz2022pretrain,   jamshidian2025blast}. While such methods are effective under homogeneous or near-homogeneous regimes, they can perform poorly even when the source and target parameters are directionally aligned but differ substantially in magnitude. This phenomenon is frequently encountered in health surveys, trial registries, and electronic health records, where covariate effects on the response variable may share signs and relative importance but differ in scale across the source and target datasets.

In that vein, recent works \citep{duan2022robust} have recognized the importance of directional similarity as an alternative notion of transferability. Angle-based transfer learning methods replace Euclidean proximity with constraints or penalties that encourage angular alignment between the source and target coefficient vectors. These approaches are appealing because a small angular deviation between the source and target parameters implies shared predictive structure across datasets even when the overall scale differs. However, existing angle-based methods \citep{duan2022robust} remain fundamentally point-wise: they rely on the direction of a single source estimator and encode directional similarity through deterministic penalties added to the target loss. As a result,  they do not readily account for uncertainty in the regression parameter estimate from the source dataset. 

Importantly, in many modern applications, the external source information available to us is not merely a point estimate of the regression parameter but an entire posterior distribution or large sample distribution of source parameters. This  motivates a shift from point-wise angular transfer to distributional angular transfer, where the goal is to encourage the direction of target parameter to be close to the directions that are assigned high probability by the source dataset. Such a perspective enables us to borrow strength in a way that is both source uncertainty-aware and robust to scale heterogeneity between the source and the target dataset.

\noindent\textbf{\underline{Our contributions.}}
\underline{First}, we introduce a framework for \emph{uncertainty-aware}, and \emph{scale-invariant} transfer learning based on directional distributional transfer, which represents source information as a distribution over directions on the unit sphere. This formulation captures uncertainty and heterogeneity in the source dataset(s) via the directional distribution of source parameters, enables alignment of the target parameter with the directions of the source regression parameter with high probability while allowing its magnitude  to be learned solely from target data.
\underline{Second}, we instantiate this idea under linear model equipped with a novel \emph{angular cone prior} that encourages scale-invariant directional alignment of  the source and the target regression parameters with high probability, and develop an efficient posterior computation scheme based on low-dimensional augmentation of the parameter space. We also demonstrate  that the framework naturally extends to multiple heterogeneous source datasets and generalized linear models.
\underline{Third}, under some mild regularity conditions, we establish posterior contraction under the proposed model and prior specification, and derive conditions under which the proposed directional distributional  transfer yields advantages over point-wise directional transfer methods. Finally, extensive numerical experiments and an application on a benchmark task demonstrate that the proposed approach consistently improves predictive performance in heterogeneous settings compared to the competing methods.

\section{Methodology}\label{method}

\subsection{Directional distribution guided transfer learning}
\begin{figure}[t]
\centering
\includegraphics[width=\linewidth]{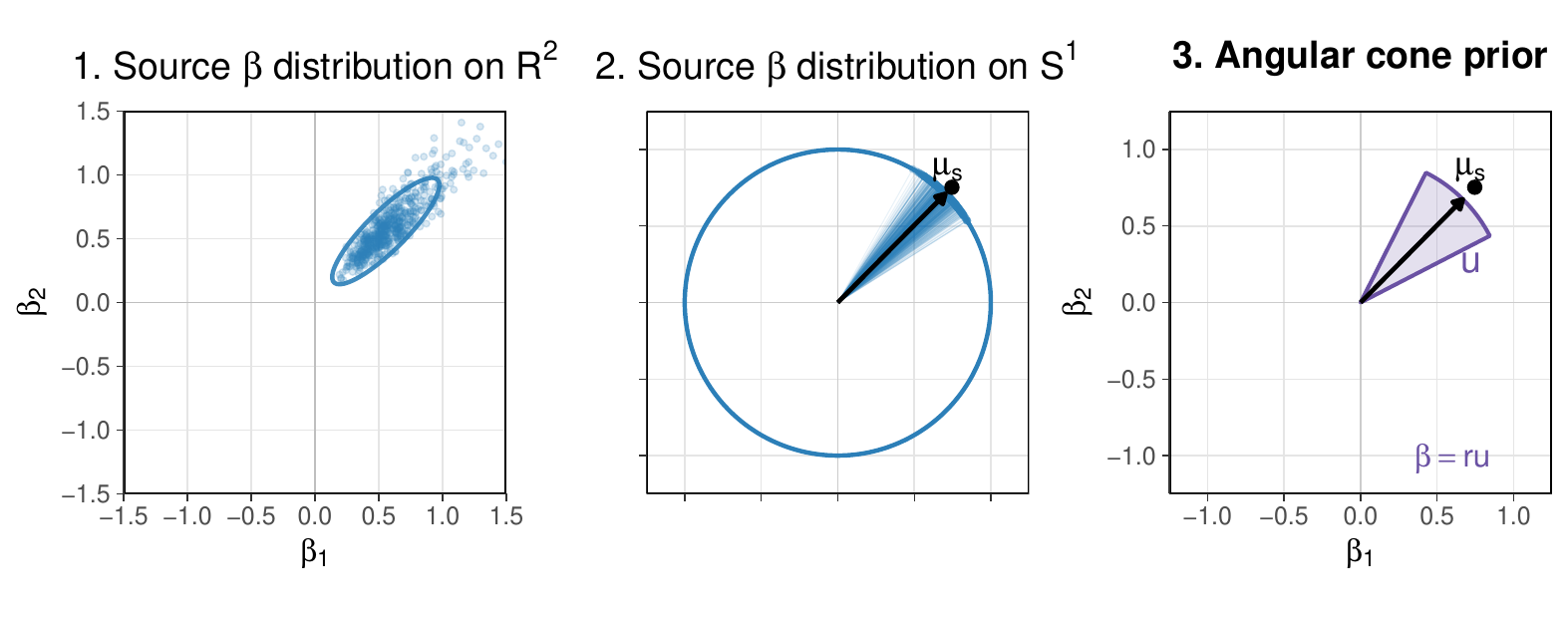}
\caption{\textbf{Uncertainty-aware and scale-invariant directional transfer learning via angular cone prior.}
(1) Source coefficient distribution in $(\beta_1,\beta_2)$ space. 
(2) Corresponding directional distribution on $\mathbb{S}^{1}$ under the mapping $T(\beta)=\beta/\|\beta\|_{2}$, with mean direction $\mu_s$. 
(3) Angular cone prior centered at $\mu_s$, encouraging directional alignment while decoupling magnitude estimation.}
\label{fig:directional_transfer_prev}
\end{figure}

We illustrate the framework in a linear regression setting to highlight its key properties without introducing unnecessary complexity; however, the framework can be readily adopted in more general models. To that end, we assume that $(Y,X)$ denote the target dataset, where $Y \in \mathbb{R}^n$ is the response and $X \in \mathbb{R}^{n \times p}$ is the covariate matrix, and consider the Gaussian linear regression model
\begin{align}
Y = X\beta + \varepsilon, 
\qquad 
\varepsilon \sim \mathcal{N}(0,\sigma^2 I_n),\quad n>>p,
\end{align}
with unknown regression coefficient $\beta \in \mathbb{R}^p$ and unknown error variance $\sigma^2>0$. External source information on $\beta$ is available in the form of a probability distribution $\pi_s$ on $\mathbb{R}^p$. This distribution may arise from a Bayesian posterior,  a large-sample Gaussian approximation of the posterior, or a bootstrap distribution. For any non-zero vector $v\in \mathbb{R}^p$, we define its direction by the map $T(v) = v/\|v\|_2 \in S^{p-1}$, where $S^{p-1}$ denotes the $p$-dimensional unit sphere. The source distribution of $\pi_s$ of $\beta$ induces a probability measure $\mu_s(\cdot)$ on  $S^{p-1}$, such that 
$$
\mu_s(A) = \pi_s\{w : T(w) \in A\}, 
\quad
A \subset S^{p-1}.
$$
The measure $\mu_s$ captures directional distribution of the source regression parameter while discarding its scale; see Figure~\ref{fig:directional_transfer_prev} for a visual representation. Transfer from source to target is thus reduced to borrowing information about the direction of $\beta$ from $\mu_s$, while solely allowing the target data to determine its magnitude. This formulation is inherently scale-invariant, and does not merely rely on the source point estimator and allows uncertainty in the source information to be propagated to the target through the distribution $\mu_s$.

We model the alignment of  $\beta$ with a candidate source direction $U \in \mathbb{S}^{p-1}$ through a geometric decomposition of the target parameter. Any coefficient vector $\beta \in \mathbb{R}^p$ can be written as
\[
\beta = aU + b, 
\qquad 
b \in \operatorname{span}(U)^\perp,
\]
where $\operatorname{span}(U)^\perp := \{v \in \mathbb{R}^p : U^\top v = 0\}$ is the orthogonal complement of the  subspace spanned by $U$, $a = \beta^\top U$ captures the component along $U$, and $b$ represents deviation orthogonal to $U$. The angular discrepancy between $\beta$ and $U$ is governed by the relative magnitude of these components, with $\tan \Theta(\beta,U) = \|b\|_2 / |a|$.
To encourage directional alignment, we place a prior that adaptively penalizes orthogonal deviation relative to the signal strength along $U$:
\begin{equation}\label{prior}
a \sim \mathcal{N}(0,\tau_a^2), 
\qquad 
b \mid a, U \sim \mathcal{N}\!\left(0,\frac{c^2}{a^2+\varepsilon}\,P_{U^\perp}\right),
\tag{2.1}
\end{equation}
where $P_{U^\perp} = I - UU^\top$ is the orthogonal projection operator onto $\operatorname{span}(U)^\perp$, $c>0$ controls angular dispersion, and $\varepsilon > 0$ is a small regularization constant that prevents degeneracy and ensures computational stability.
We term the construction in Equation \ref{prior} as an \emph{angular cone prior}, under which $\beta$ concentrates around the ray spanned by $U$, with the degree of concentration adapting to $|a|$. Larger values of $|a|$ correspond to stronger alignment and lead to tighter concentration around $U$, while smaller values allow greater angular spread. 
The hierarchical specification is completed by placing an inverse-gamma prior on the noise variance, i.e., $\sigma^2 \sim \mathrm{IG}(\alpha_0,\beta_0)$.

We further extend the proposed framework to multiple heterogeneous sources and generalized linear models; details are provided in the supplementary material.

\subsection{Operating characteristics of the angular cone prior}\label{sec:properties}

The angular cone prior induces a structured geometric shrinkage, with three key properties that together characterize its behavior: explicit control of angular uncertainty, signal-adaptive shrinkage, and scale-invariant transfer.

\begin{enumerate}
    \item The prior admits an exact finite-sample characterization of angular concentration. Conditional on $(a,U)$, we have $\frac{a^2+\varepsilon}{c^2}\,\|b\|_2^2 \sim \chi^2_{p-1}$. Consequently, the angular deviation $\Theta(\beta,U)$ satisfies $\mathbb{P}[\Theta(\beta,U)\le t \mid a,U] = F_{\chi^2_{p-1}}\!\left(\frac{a^2(a^2+\varepsilon)}{c^2}\tan^2 t\right)$ for $t\in(0,\pi/2)$.
The $(1-\delta)$ angular credible region is therefore given by $\Theta(\beta,U)\le \arctan\!\left(\sqrt{\frac{c^2}{a^2(a^2+\varepsilon)}\,F^{-1}_{\chi^2_{p-1}}(1-\delta)}\right)$, which shrinks monotonically in $|a|$, yielding tighter concentration as alignment strengthens.

\item  The prior exhibits signal-adaptive directional shrinkage. The expected squared angular deviation satisfies
\begin{equation}
\mathbb{E}\{\tan^2\Theta(\beta,U)\mid a,U\}
=
\frac{(p-1)c^2}{a^2(a^2+\varepsilon)},
\tag{2.3}
\end{equation}
which decreases monotonically in $|a|$. Thus, strong alignment induces concentrated borrowing along $U$, while weak alignment allows broader exploration of directions. This adaptivity arises intrinsically from the prior construction and does not require additional tuning.

\item The induced transfer mechanism is scale-invariant. For any $s>0$, replacing a source coefficient $w$ by $sw$ leaves the directional distribution $\mu_s$ unchanged, and hence does not affect the induced prior or posterior on $\beta$. 
In particular, when source and target parameters are directionally aligned but differ substantially in norm, the prior avoids the bias inherent in Euclidean shrinkage.
\end{enumerate}

\begin{figure}
\centering
\fbox{%
\begin{minipage}{0.96\linewidth}
\small
\textbf{Algorithm 1} \quad Directional transfer via angular cone prior
\hrule

\textbf{Input:} Data $(Y,X)$, directional prior $\mu_s$,\\
\textbf{Output:} Posterior samples $\{\beta^{(m)},\sigma^{2\,(m)}\}_{m=1}^M$
\hrule
\textbf{Initialize:} $(a,\theta,U,\sigma^2)$
\begin{enumerate}
    \item \textbf{Projection:} Compute $x_U = XU$ and $Z_U = XP_{U^\perp}$.
    \item \textbf{Orthogonal update:} Sample $\theta$ from its Gaussian full conditional.
    \item \textbf{Alignment update:} Sample $a$ via slice sampling.
    \item \textbf{Noise update:} Sample $\sigma^2$ from its inverse-gamma full conditional.
    \item \textbf{Direction update:} 
    (a) if $\mu_s$ is discrete, sample $U$ from its finite support;
    (b) if $\mu_s$ is continuous, update $U$ on $\mathbb{S}^{p-1}$ via Metropolis--Hastings or Hamiltonian Monte Carlo.
    \item \textbf{Reconstruction:} Set $\beta = aU + Q_U\theta$.
\end{enumerate}
\end{minipage}%
}
\end{figure}

\subsection{Posterior sampler and geometry-aware computational strategies}\label{sec:comp}
Posterior inference is performed via a Gibbs sampler summarized in Algorithm 1. The computational strategy is built on the orthogonal reparameterization $\beta = aU + Q_U\theta$, where $U \in \mathbb{S}^{p-1}$, $a \in \mathbb{R}$ controls alignment with $U$, and $\theta \in \mathbb{R}^{p-1}$ captures orthogonal deviation.
\vspace{2mm}

\textbf{\underline{Step 1 (Projection).}}
We compute $x_U = XU$ and $Z_U = XQ_U$, where $U \in \mathbb{S}^{p-1}$ is a unit direction and $Q_U \in \mathbb{R}^{p \times (p-1)}$ is an orthonormal basis for the subspace orthogonal to $U$, i.e., $Q_U^\top Q_U = I$ and $Q_U Q_U^\top = P_{U^\perp}$. Equivalently, the orthogonal projection can be expressed using the projection operator $P_{U^\perp} = I - UU^\top$, so that $Z_U = X P_{U^\perp}$ without explicitly constructing $Q_U$.
\vspace{2mm}

\textbf{\underline{Step 2 (Orthogonal update).}}
Conditioned on $(a,U,\sigma^2)$, $\theta$ follows Gaussian distribution with precision $\Lambda_\theta = \sigma^{-2} Z_U^\top Z_U + (a^2+\varepsilon)c^{-2} I$ and mean $m_\theta = \Lambda_\theta^{-1}\sigma^{-2} Z_U^\top (Y - a x_U)$. Sampling $\theta\mid a, U,\sigma^2$ is therefore equivalent to solving the linear system
$\Lambda_\theta \, \theta = \sigma^{-2} Z_U^\top (Y - a x_U) + \eta$, where $\eta \sim \mathcal{N}(0,\Lambda_\theta)$.
This is efficiently implemented via a Cholesky factorization $\Lambda_\theta = LL^\top$ by solving two triangular systems. In practice, one computes $v = \sigma^{-2} Z_U^\top (Y - a x_U)$, solves $L y = v$, then $L^\top \theta = y + \xi$ with $\xi \sim \mathcal{N}(0,I)$.
\vspace{2mm}

\textbf{\underline{Step 3 (Alignment update).}}
The scalar $a \mid \theta, U, \sigma^2$ is updated via slice sampling \cite{neal2003slice} from its one-dimensional conditional density with log-density proportional to
\[
-\|Y-a x_U - Z_U\theta\|_2^2/(2\sigma^2) - a^2/(2\tau_a^2) + \frac{p-1}{2}\log(a^2+\varepsilon) - \frac{a^2+\varepsilon}{2c^2}\|\theta\|_2^2.
\]
Slice sampling introduces an auxiliary variable $u \sim \mathrm{Unif}(0,f(a))$, where $f(a)$ denotes the unnormalized conditional density, and samples $a$ uniformly from the slice $\mathcal{S} = \{a : \log f(a) \ge \log u\}$. In practice, this is implemented via a stepping-out procedure to construct an interval containing $\mathcal{S}$, followed by shrinkage steps to obtain a valid draw. 
\vspace{2mm}

\textbf{\underline{Step 4 (Noise variance update).}}
The variance parameter $\sigma^2\mid \theta, a, U$ is sampled from its inverse-gamma conditional distribution. 
\vspace{2mm}

\textbf{\underline{Step 5 (Direction update).}}
The latent direction $U\mid \theta, a, \sigma$ is updated on $\mathbb{S}^{p-1}$, with the sampling strategy that depends on the nature of $\mu_s$. If $\mu_s$ is discrete, $U\mid \theta, a, \sigma$ is sampled from a finite set with weights proportional to $\exp\{-\|Y-a x_U - Z_U\theta\|_2^2/(2\sigma^2)\},$ yielding a finite-mixture update. If $\mu_s$ is continuous, $U\mid \theta, a, \sigma$ is updated via a Metropolis step on the sphere. 
\vspace{2mm}

Random-walk Metropolis  can mix poorly when the posterior is concentrated or multi-modal; in this regime, we employ Hamiltonian Monte Carlo (HMC, \cite{neal2011mcmc, byrne2013geodesic}) on $\mathbb{S}^{p-1}$.
Specifically, HMC introduces an auxiliary momentum variable in the tangent space of the sphere and simulates Hamiltonian dynamics constrained to $\mathbb{S}^{p-1}$ via geodesic flows. Gradients of the log-posterior are computed using the projection operator $P_{U^\perp} = I - UU^\top$, ensuring that updates remain on the manifold while avoiding explicit construction of $Q_U$.
\vspace{2mm}

\textbf{\underline{Step 6 (Reconstruction).}}
Finally, we set $\beta = aU + Q_U\theta$. This recombines the aligned and orthogonal components into the full parameter vector.

The most substantial computational cost arises from computing projections in Step~1. This is mitigated by caching $XU$ for discrete $\mu_s$, using projection operators to avoid recomputing $Q_U$, and exploiting low-rank updates under local proposals for $U$. Each iteration therefore alternates between Euclidean updates in the orthogonal subspace (Steps~2--4) and geometric updates on the sphere (Step~5).  Algorithmic details and implementation refinements are deferred to the supplementary material.

\section{Theoretical analysis}\label{theory}
\subsection{Posterior consistency results}\label{contraction}
\label{sec:contraction}

We study posterior concentration \citep{ghosal2000convergence, ghosal2007convergence, ghosal2017fundamentals} under the linear model $Y = X\beta_0 + \varepsilon,\
\varepsilon \sim \mathcal N(0,\sigma^2 I_n)$ equipped with the proposed angular cone prior. We recall that $T(v)=v/\|v\|_2$ and define the true direction $U_0^\star=T(\beta_0)$. The prior on $\beta$ is defined hierarchically by $U \sim \mu_s, \beta = aU + b, b\in \mathrm{span}(U)^\perp,$
with $a \sim \mathcal N(0,\tau_a^2),\ b \mid a,U \sim \mathcal N\!\left(0,\frac{c^2}{a^2+\varepsilon}P_{U^\perp}\right)$, where $P_{U^\perp}=I-UU^\top$, $c>0$ and $\varepsilon>0$. Let $\Pi(\cdot\mid Y, X)$ denote the corresponding posterior distribution. We impose the following assumptions.

\begin{assumption}[Design regularity]
\label{ass:design}
There exist constants $0<\kappa_-<\kappa_+<\infty$ such that
$\kappa_- \le \lambda_{\min}(\frac{X^\top X}{n})
\le \lambda_{\max}\!(\frac{X^\top X}{n})
\le \kappa_+$,
uniformly in $n$.
\end{assumption}

\begin{assumption}[Non-degeneracy]
\label{ass:a_nonzero}
The true coefficient satisfies $\|\beta_0\|_2<\infty$.
\end{assumption}

Assumption~\ref{ass:design} is standard and
ensures local quadratic behavior of the log-likelihood.
Assumption~\ref{ass:a_nonzero} guarantees that the true parameter has finite Euclidean norm, ensuring well-defined contraction in the reparameterization $\beta = aU + Q_U\theta$.

\begin{theorem}[Directional posterior contraction]
\label{thm:direction_contraction}
Assume in addition that $\|\beta_0\|_2\ge c_0>0$.
Then there exists $M>0$ such that
$\Pi(\Theta(T(\beta),U_0^\star)>
M \frac{1}{\|\beta_0\|_2}\sqrt{\frac{p}{n}}
\mid Y,X
)\;\xrightarrow[n\to\infty]{P_{\beta_0}} 0$.
\end{theorem}
The rate reflects the geometry of the sphere: $\Theta(T(\beta),U_0^\star) \asymp \|\beta-\beta_0\|_2 / \|\beta_0\|_2$. In particular, for a fixed Euclidean error $\|\beta-\beta_0\|_2$, the induced angular error is inversely proportional to $\|\beta_0\|_2$, so smaller signal strength leads to larger angular uncertainty.

Next, we recall that the main feature of the proposed approach is that transfer occurs through the directional distribution $\mu_s$. To formalize when this yields sharper concentration, we first impose a mild compatibility condition.


\begin{assumption}[Localized source directional information]
\label{ass:source_localization}
Suppose that $\beta_0\neq 0$. Let $\mu_{s,n}$ denote the directional
distribution induced by the available source information, and define
$U_0^\star=T(\beta_0)=\beta_0/\|\beta_0\|_2$. Let
$\epsilon_n=\sqrt{p/n}$, where $p$ is fixed. There exists a constant
$C_s>0$ such that, for all sufficiently large $n$,
$$
\mu_{s,n}
\left(
\left\{
U\in\mathbb S^{p-1}:
\Theta(U,U_0^\star)\le C_s\epsilon_n
\right\}
\right)
=1.
$$
\end{assumption}
Assumption~\ref{ass:source_localization} states the localization
property implicitly required for enhanced directional concentration under transfer. In particular,  we assume that
the external source information is substantially more precise than the target
information and the residual source--target discrepancy is predominantly in
magnitude rather than direction. In such settings, the directional uncertainty
in the source coefficient is expected to be smaller than, or at most of the
same order as, the target directional estimation uncertainty.
More formally, let $U_{s,n}^\star$ denote the main direction of the source
distribution. By the triangle inequality for the geodesic distance on the
sphere,
$$
\Theta(U,U_0^\star)
\le
\Theta(U,U_{s,n}^\star)
+
\Theta(U_{s,n}^\star,U_0^\star).
$$
Therefore, Assumption~\ref{ass:source_localization} is satisfied whenever
$$
\sup_{U\in\operatorname{supp}(\mu_{s,n})}
\Theta(U,U_{s,n}^\star)
\le C_1\epsilon_n,\quad\text{and}\quad
\Theta(U_{s,n}^\star,U_0^\star)
\le C_2\epsilon_n
$$
for constants $C_1,C_2>0$. The triangle inequality then gives
$
\sup_{U\in\operatorname{supp}(\mu_{s,n})}
\Theta(U,U_0^\star)
\le
(C_1+C_2)\epsilon_n,
$
so Assumption~\ref{ass:source_localization} holds with
$C_s=C_1+C_2$.

This condition does not require the source and target coefficient vectors to
be identical or exactly proportional. Their directions may differ at finite
sample sizes, provided that this discrepancy is no larger than the target
directional estimation scale. This is the regime in which directional
borrowing is theoretically expected to be beneficial: the source supplies
directional information that is at least as accurate as what can be learned
from the target sample alone, while the target data remain responsible for
estimating the magnitude and refining the target coefficient.

From a practical perspective, the source directional distribution used by
DUST is typically represented by a finite empirical collection of posterior
or bootstrap draws,
$
\mu_{s,n}
=
\frac{1}{K_n}
\sum_{k=1}^{K_n}
\delta_{U_{s,n}^{(k)}}.
$
Consequently, the probability-one support condition in
Assumption~\ref{ass:source_localization} is a condition on the finite set of
directions actually supplied to the transfer procedure, rather than on the
unbounded support of an idealized continuous Gaussian approximation. When the
source dataset is sufficiently informative, the posterior or bootstrap draws
concentrate tightly around their central direction, and the empirical
directional distribution is contained in the prescribed shrinking cap with
probability tending to one. In applications, the concentration of the source
information can be examined directly through the empirical angular dispersion
of the source draws. The assumption therefore formalizes a practically
relevant regime in which the external directional information is precise and
locally compatible with the target task.

\begin{theorem}[Enhanced directional concentration under transfer]
\label{thm:transfer_gain}
Let $\epsilon_n=\sqrt{p/n}$, where $p$ is fixed. Suppose that
Assumptions~\ref{ass:design}, \ref{ass:a_nonzero}, and
\ref{ass:source_localization} hold. Then, for every constant $M>C_s$,
$$
\Pi\left(
\Theta(U,U_0^\star)>M\epsilon_n
\,\middle|\,
Y,X
\right)
=
0
$$
for all sufficiently large $n$.
If, in addition, $\|\beta_0\|_2\ge c_0>0$ and the posterior for $\beta$
contracts around $\beta_0$ at rate $\epsilon_n$, then, for every sufficiently
slowly diverging sequence $M_n\to\infty$,
$$
\Pi\left(
\Theta\{T(\beta),U_0^\star\}>M_n\epsilon_n
\,\middle|\,
Y,X
\right)
\xrightarrow[n\to\infty]{P_{\beta_0}}0.
$$
\end{theorem}

\subsection{Relative gain from distributional directional transfer}\label{sec:relative_gain}

Next, we formalize the advantage of distributional directional transfer by characterizing prediction risk in terms of directional alignment.
We work under the following assumptions.

\begin{assumption}
\label{assump:model}
The target data follow the linear model
$Y = X^\top \beta_0 + \varepsilon$,
where $\mathbb{E}(X)=0$, $\mathrm{Cov}(X)=I_p$, and
$\mathbb{E}(\varepsilon \mid X)=0$, $\mathrm{Var}(\varepsilon \mid X)=\sigma^2$.
Let $\beta_0 = b_0 u_0$ with $b_0=\|\beta_0\|_2$ and $u_0 \in \mathbb{S}^{p-1}$.
\end{assumption}

\begin{assumption}
\label{assump:source}
External source information is summarized by a probability distribution $G$
on the unit sphere $\mathbb{S}^{p-1}$. Let
$m_G = \mathbb{E}_G(U)$ and define $u_{\mathrm{pt}} = m_G/\|m_G\|_2$,
provided $m_G \neq 0$.
\end{assumption}
We first characterize the risk of point-wise directional transfer.

\begin{theorem}[Risk under point-wise directional transfer]
\label{thm:relative_gain}
For any direction $u \in \mathbb{S}^{p-1}$, consider the optimally scaled predictor $a^\star(u)u$, where $a^\star(u) = \arg\min_{a \in \mathbb{R}} \mathbb{E}(Y - a X^\top u)^2$. Then $a^\star(u) = \beta_0^\top u$, and the excess prediction risk satisfies
$$R(u) = \mathbb{E}(Y - X^\top a^\star(u)u)^2 - \sigma^2 =
\|\beta_0\|_2^2\bigl(1-(u_0^\top u)^2\bigr).$$
\end{theorem}

Theorem~\ref{thm:relative_gain} shows that the risk of point-wise directional transfer depends only on squared cosine alignment with the true direction. We next consider distributional directional transfer, defined as the procedure that draws $U \sim G$ and uses the optimally scaled predictor $a^\star(U)U$. The resulting  risk is
$R(G):=\mathbb{E}_{U \sim G}\big[\mathbb{E}(Y - X^\top a^\star(U)U)^2\big]- \sigma^2$.


\begin{figure}[t]
\centering
\includegraphics[width=\linewidth]{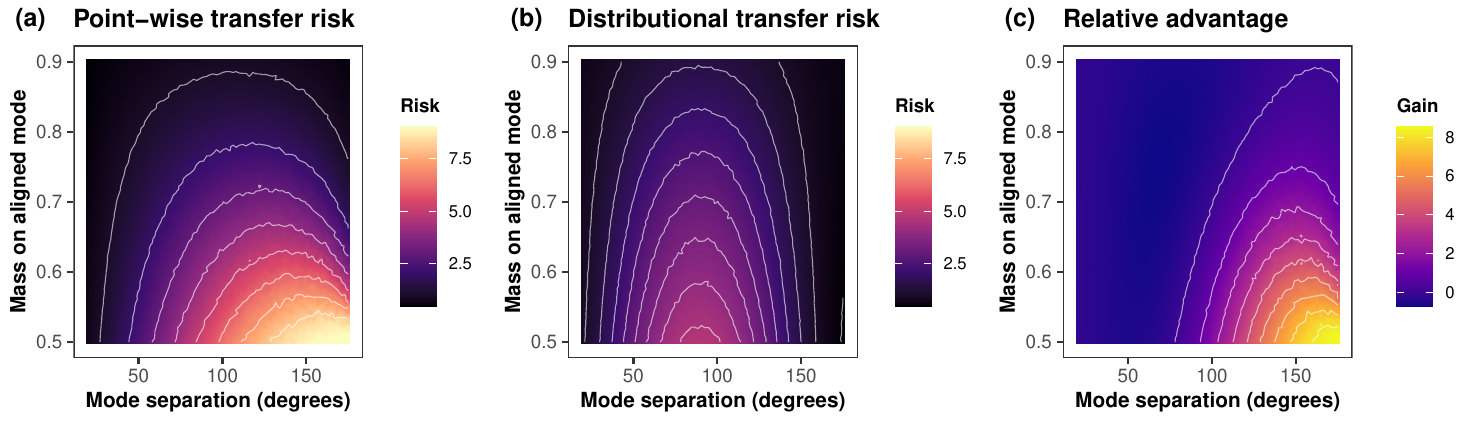}
\caption{\textbf{Prediction risk under heterogeneous directional source information.}
Assume that the directional distribution of the source coefficient is bi-modal. The horizontal axis varies the angular separation between source directional components, while the vertical axis varies the proportion of source mass assigned to the direction aligned with the target signal. 
\emph{(a)} Risk of point-wise transfer, which increases as source directions become more separated due to averaging across conflicting signals. 
\emph{(b)} Risk of the proposed distributional transfer estimator, which remains comparatively stable by aggregating over multiple plausible source directions. 
\emph{(c)} Relative gain of DUST over point-wise transfer, showing the largest improvement in regimes with heterogeneous yet partially informative source structure. Lower values indicate better predictive performance.}
\label{fig:relative_gain_surface}
\end{figure}

\begin{theorem}[Risk under distributional directional transfer]
\label{thm:dist_gain}
The excess mixture risk satisfies
$$R(G) = \|\beta_0\|_2^2 \bigl(1 - \mathbb{E}_G[(u_0^\top U)^2]\bigr).$$
\end{theorem}
Comparing Theorems~\ref{thm:relative_gain} and~\ref{thm:dist_gain}, the gain of distributional directional transfer over point-wise transfer is
\[
\Delta_{\mathrm{gain}}
=
R(u_{\mathrm{pt}}) - R(G)
=
\|\beta_0\|_2^2
\Bigl(
\mathbb{E}_G[(u_0^\top U)^2]
-
(u_0^\top u_{\mathrm{pt}})^2
\Bigr).
\]

Jensen's inequality and the normalization of $u_{\mathrm{pt}} = m_G/\|m_G\|_2$, we have
$$(u_0^\top u_{\mathrm{pt}})^2=\frac{(u_0^\top m_G)^2}{\|m_G\|_2^2}\le\mathbb{E}_G[(u_0^\top U)^2],$$
with equality if and only if $u_0^\top U$ is almost surely constant under $G$. Consequently, $\Delta_{\mathrm{gain}} \ge 0$, with strict inequality whenever $G$ is not degenerate.


Finally, the oracle rule $u_{\mathrm{or}} \in \arg\max_{u \in \mathrm{supp}(G)} |u_0^\top u|$ achieves risk
$R(u_{\mathrm{or}})= \|\beta_0\|_2^2\bigl(1-c_{\mathrm{or}}^2\bigr),
c_{\mathrm{or}} = \sup_{u \in \mathrm{supp}(G)} |u_0^\top u|$,
and hence
$R(u_{\mathrm{or}}) \le R(G) \le R(u_{\mathrm{pt}})$.

\section{Numerical examples}\label{exp}
In all the numerical examples, we compare four estimators: \texttt{target-only}, \texttt{Euclidean transfer}, \texttt{angular point transfer}, and  the proposed \texttt{directional distributional transfer (DUST)}.  All experiments were conducted on a standard a Dell XPS 15 laptop with an Intel Core i7 processor with 8 cores and 32 GB RAM.

\subsection{Simulation studies}\label{sec:sim}

\noindent\textbf{\underline{A. Linear regression.}}
We consider the linear regression model $Y = X\beta_0 + \varepsilon$ with $\varepsilon \sim \mathcal{N}(0, I_n)$, where $p \in \{10,20,30,40\}$, sample size $n \in \{100,200,300,400,500\}$, and rows of $X \in \mathbb{R}^{n\times p}$ are generated as $X_i \sim \mathcal{N}(0,\Sigma)$ with $\Sigma_{jk} = 0.8^{|j-k|}$. The true regression parameter is $\beta_0 = s u_0$ with $u_0 \sim \mathrm{Unif}(\mathbb{S}^{p-1})$ and $s=2$, isolating its direction from scale. Performance is evaluated using test mean squared error averaged over Monte Carlo repetitions under identical training/test splits. We consider two different heterogeneous source data generating mechanisms:

\textbf{(1) Skewed uni-modal source dataset.}
We construct a discrete, uni-modal yet asymmetric directional distribution of the source parameter, in which the bulk of directions concentrate near the target (deviation $\leq20^\circ$), while a small fraction forms a skewed shoulder along a common drift direction (deviation between $20^\circ$ and $35^\circ$); the details are deferred to the supplement. This induces a mild scale–direction mismatch between target and source parameters, and introduces asymmetry in the source parameter distribution, rendering point summaries of source distribution slightly misleading.
Table \ref{tab:linear_unimodal_compact} demonstrates the advantage of the proposed DUST method across all $(n,p)$ configurations. Euclidean transfer fares worse compared to DUST due to its sensitivity to scale heterogeneity between source and target regression parameters. Although, angular transfer is capable of handing scale mismatch, the asymmetric source distribution leads to systematic directional bias in the angular estimator and degrades its performance  relative to even the target-only estimator in several regimes.


\textbf{(2)  Bimodal source dataset.}
We consider a bimodal directional distribution of source parameter, i.e., $\mu_s = 0.6\,\delta_{u_1} + 0.4\,\delta_{u_2}$, where $u_1$ is close to $u_0$ ($\approx 15^\circ$) and $u_2$ is misaligned ($\approx 60^\circ$). 
Table~\ref{tab:linear_bimodal_compact} shows that the proposed DUST method consistently attains the lowest error across $(n,p)$. Gains are largest for small $n$, reflecting improved variance reduction via directional borrowing. The advantage of the proposed method remains stable with $p$.  Point-wise angular transfer degrades under bimodality due to alignment to a single, potentially misaligned direction, while Euclidean transfer falls prey to the source-target scale mismatch.  

\begin{table}[t]
\tiny
\setlength{\tabcolsep}{3.5pt}

\begin{minipage}{0.49\linewidth}
\centering
\caption{\textbf{Linear regression (directional transfer under unimodal heterogeneous source structure).}
Mean test MSE with standard error in parentheses (lower is better).}
\label{tab:linear_unimodal_compact}
\begin{tabular}{cccccc}
\toprule
$n$ & $p$ & Target-only & Euclidean & Angular & DUST \\
\midrule
100 & 10 & 0.117 (0.009) & 0.107 (0.007) & 0.112 (0.012) & \textbf{0.012 (0.002)} \\
100 & 20 & 0.250 (0.010) & 0.165 (0.007) & 0.096 (0.007) & \textbf{0.012 (0.002)} \\
100 & 30 & 0.454 (0.018) & 0.263 (0.010) & 0.109 (0.005) & \textbf{0.011 (0.002)} \\
100 & 40 & 0.677 (0.034) & 0.386 (0.019) & 0.118 (0.010) & \textbf{0.013 (0.002)} \\
\midrule
200 & 10 & 0.046 (0.003) & 0.071 (0.005) & 0.089 (0.010) & \textbf{0.007 (0.001)} \\
200 & 20 & 0.103 (0.005) & 0.093 (0.004) & 0.119 (0.010) & \textbf{0.005 (0.001)} \\
200 & 30 & 0.187 (0.009) & 0.136 (0.008) & 0.109 (0.009) & \textbf{0.005 (0.001)} \\
200 & 40 & 0.247 (0.008) & 0.165 (0.006) & 0.114 (0.007) & \textbf{0.006 (0.001)} \\
\midrule
300 & 10 & 0.035 (0.002) & 0.059 (0.004) & 0.092 (0.009) & \textbf{0.005 (0.001)} \\
300 & 20 & 0.070 (0.003) & 0.075 (0.004) & 0.092 (0.007) & \textbf{0.004 (0.001)} \\
300 & 30 & 0.109 (0.003) & 0.105 (0.005) & 0.123 (0.009) & \textbf{0.004 (0.001)} \\
300 & 40 & 0.165 (0.006) & 0.126 (0.005) & 0.116 (0.008) & \textbf{0.004 (0.001)} \\
\midrule
400 & 10 & 0.027 (0.001) & 0.055 (0.004) & 0.101 (0.013) & \textbf{0.004 (0.001)} \\
400 & 20 & 0.052 (0.003) & 0.069 (0.004) & 0.132 (0.009) & \textbf{0.003 (0.000)} \\
400 & 30 & 0.082 (0.003) & 0.089 (0.004) & 0.106 (0.007) & \textbf{0.003 (0.001)} \\
400 & 40 & 0.114 (0.004) & 0.111 (0.005) & 0.122 (0.009) & \textbf{0.003 (0.000)} \\
\midrule
500 & 10 & 0.019 (0.001) & 0.059 (0.004) & 0.090 (0.009) & \textbf{0.002 (0.000)} \\
500 & 20 & 0.041 (0.002) & 0.066 (0.004) & 0.108 (0.007) & \textbf{0.003 (0.000)} \\
500 & 30 & 0.067 (0.003) & 0.072 (0.003) & 0.119 (0.008) & \textbf{0.002 (0.001)} \\
500 & 40 & 0.089 (0.003) & 0.091 (0.004) & 0.105 (0.006) & \textbf{0.002 (0.000)} \\
\bottomrule
\end{tabular}
\end{minipage}
\hfill
\begin{minipage}{0.49\linewidth}
\centering
\caption{\textbf{Linear model (directional transfer under bimodal source heterogeneity).}
Mean test MSE with standard error in parentheses (lower is better).}
\label{tab:linear_bimodal_compact}
\begin{tabular}{cccccc}
\toprule
$n$ & $p$ & Target-only & Euclidean & Angular & DUST \\
\midrule

100 & 10 & 1.114 (0.007) & 1.133 (0.009) & 1.289 (0.019) & \textbf{1.019 (0.002)} \\
100 & 20 & 1.249 (0.012) & 1.226 (0.010) & 1.361 (0.019) & \textbf{1.013 (0.002)} \\
100 & 30 & 1.416 (0.016) & 1.333 (0.012) & 1.350 (0.015) & \textbf{1.016 (0.002)} \\
100 & 40 & 1.688 (0.041) & 1.477 (0.023) & 1.352 (0.013) & \textbf{1.014 (0.002)} \\
\midrule
200 & 10 & 1.058 (0.003) & 1.066 (0.003) & 1.377 (0.029) & \textbf{1.013 (0.001)} \\
200 & 20 & 1.113 (0.005) & 1.109 (0.004) & 1.371 (0.024) & \textbf{1.011 (0.001)} \\
200 & 30 & 1.178 (0.007) & 1.159 (0.006) & 1.328 (0.015) & \textbf{1.010 (0.001)} \\
200 & 40 & 1.241 (0.007) & 1.212 (0.007) & 1.344 (0.014) & \textbf{1.011 (0.001)} \\
\midrule
300 & 10 & 1.035 (0.002) & 1.035 (0.002) & 1.327 (0.022) & \textbf{1.008 (0.001)} \\
300 & 20 & 1.073 (0.004) & 1.070 (0.004) & 1.308 (0.017) & \textbf{1.009 (0.001)} \\
300 & 30 & 1.108 (0.004) & 1.103 (0.004) & 1.331 (0.016) & \textbf{1.010 (0.001)} \\
300 & 40 & 1.160 (0.005) & 1.146 (0.004) & 1.349 (0.015) & \textbf{1.010 (0.002)} \\
\midrule
400 & 10 & 1.028 (0.001) & 1.029 (0.002) & 1.346 (0.028) & \textbf{1.007 (0.001)} \\
400 & 20 & 1.055 (0.003) & 1.054 (0.003) & 1.339 (0.019) & \textbf{1.007 (0.001)} \\
400 & 30 & 1.083 (0.003) & 1.079 (0.003) & 1.331 (0.015) & \textbf{1.007 (0.001)} \\
400 & 40 & 1.118 (0.003) & 1.111 (0.003) & 1.342 (0.013) & \textbf{1.006 (0.001)} \\
\midrule
500 & 10 & 1.019 (0.001) & 1.020 (0.001) & 1.311 (0.026) & \textbf{1.005 (0.001)} \\
500 & 20 & 1.043 (0.002) & 1.043 (0.003) & 1.369 (0.025) & \textbf{1.006 (0.001)} \\
500 & 30 & 1.063 (0.002) & 1.060 (0.002) & 1.313 (0.013) & \textbf{1.005 (0.001)} \\
500 & 40 & 1.086 (0.003) & 1.083 (0.003) & 1.325 (0.016) & \textbf{1.007 (0.001)} \\
\bottomrule
\end{tabular}
\end{minipage}

\end{table}

\begin{table}[t]
\tiny
\setlength{\tabcolsep}{3.5pt}

\begin{minipage}{0.49\linewidth}
\centering
\caption{\textbf{Logistic regression (directional transfer under unimodal heterogeneous source structure).}
Mean test log-loss with standard error in parentheses (lower is better).}
\label{tab:logistic_unimodal_compact}
\begin{tabular}{cccccc}
\toprule
$n$ & $p$ & Target-only & Euclidean & Angular & DUST \\
\midrule
100 & 10 & 0.479 (0.017) & 0.396 (0.011) & 0.425 (0.014) & \textbf{0.387 (0.012)} \\
100 & 20 & 0.855 (0.066) & 0.421 (0.009) & 0.431 (0.010) & \textbf{0.403 (0.009)} \\
100 & 30 & 2.166 (0.118) & 0.434 (0.010) & 0.425 (0.009) & \textbf{0.395 (0.008)} \\
100 & 40 & 2.590 (0.098) & 0.432 (0.008) & 0.423 (0.008) & \textbf{0.389 (0.007)} \\
\midrule
200 & 10 & 0.433 (0.010) & 0.421 (0.009) & 0.457 (0.011) & \textbf{0.413 (0.009)} \\
200 & 20 & 0.458 (0.011) & 0.400 (0.010) & 0.417 (0.011) & \textbf{0.379 (0.009)} \\
200 & 30 & 0.543 (0.019) & 0.398 (0.009) & 0.407 (0.009) & \textbf{0.372 (0.008)} \\
200 & 40 & 0.886 (0.076) & 0.417 (0.008) & 0.414 (0.009) & \textbf{0.377 (0.007)} \\
\midrule
300 & 10 & 0.406 (0.011) & 0.400 (0.012) & 0.427 (0.013) & \textbf{0.393 (0.011)} \\
300 & 20 & 0.423 (0.010) & 0.398 (0.010) & 0.419 (0.012) & \textbf{0.384 (0.010)} \\
300 & 30 & 0.454 (0.008) & 0.399 (0.008) & 0.414 (0.008) & \textbf{0.378 (0.007)} \\
300 & 40 & 0.501 (0.008) & 0.406 (0.007) & 0.413 (0.008) & \textbf{0.373 (0.007)} \\
\midrule
400 & 10 & 0.400 (0.011) & 0.393 (0.011) & 0.428 (0.013) & \textbf{0.388 (0.011)} \\
400 & 20 & 0.398 (0.008) & 0.386 (0.008) & 0.408 (0.009) & \textbf{0.372 (0.008)} \\
400 & 30 & 0.431 (0.007) & 0.404 (0.008) & 0.427 (0.009) & \textbf{0.387 (0.008)} \\
400 & 40 & 0.452 (0.007) & 0.402 (0.006) & 0.417 (0.006) & \textbf{0.379 (0.006)} \\
\midrule
500 & 10 & 0.389 (0.011) & 0.385 (0.011) & 0.415 (0.013) & \textbf{0.380 (0.011)} \\
500 & 20 & 0.390 (0.008) & 0.381 (0.008) & 0.406 (0.009) & \textbf{0.371 (0.008)} \\
500 & 30 & 0.402 (0.008) & 0.384 (0.008) & 0.403 (0.009) & \textbf{0.367 (0.008)} \\
500 & 40 & 0.419 (0.007) & 0.391 (0.008) & 0.406 (0.008) & \textbf{0.370 (0.008)} \\
\bottomrule
\end{tabular}
\end{minipage}
\hfill
\begin{minipage}{0.49\linewidth}
\centering
\caption{\textbf{Logistic regression (directional transfer under bimodal source heterogeneity).}
Mean test log-loss with standard error in parentheses (lower is better).}
\label{tab:logistic_bimodal_compact}
\begin{tabular}{cccccc}
\toprule
$n$ & $p$ & Target-only & Euclidean & Angular & DUST \\
\midrule
100 & 10 & 0.506 (0.033) & 0.402 (0.012) & 0.398 (0.012) & \textbf{0.383 (0.011)} \\
100 & 20 & 0.905 (0.067) & 0.430 (0.011) & 0.412 (0.011) & \textbf{0.398 (0.010)} \\
100 & 30 & 2.104 (0.147) & 0.421 (0.009) & 0.401 (0.010) & \textbf{0.379 (0.009)} \\
100 & 40 & 3.030 (0.131) & 0.441 (0.010) & 0.405 (0.010) & \textbf{0.391 (0.009)} \\
\midrule
200 & 10 & 0.431 (0.011) & 0.408 (0.011) & 0.409 (0.012) & \textbf{0.396 (0.011)} \\
200 & 20 & 0.479 (0.012) & 0.413 (0.012) & 0.406 (0.013) & \textbf{0.388 (0.012)} \\
200 & 30 & 0.590 (0.040) & 0.416 (0.010) & 0.402 (0.010) & \textbf{0.387 (0.009)} \\
200 & 40 & 0.993 (0.111) & 0.411 (0.009) & 0.392 (0.009) & \textbf{0.371 (0.008)} \\
\midrule
300 & 10 & 0.392 (0.013) & 0.385 (0.013) & 0.391 (0.014) & \textbf{0.377 (0.013)} \\
300 & 20 & 0.432 (0.010) & 0.404 (0.011) & 0.404 (0.011) & \textbf{0.386 (0.010)} \\
300 & 30 & 0.461 (0.009) & 0.408 (0.009) & 0.399 (0.009) & \textbf{0.384 (0.009)} \\
300 & 40 & 0.497 (0.010) & 0.402 (0.009) & 0.389 (0.010) & \textbf{0.372 (0.009)} \\
\midrule
400 & 10 & 0.419 (0.012) & 0.412 (0.012) & 0.419 (0.013) & \textbf{0.404 (0.012)} \\
400 & 20 & 0.430 (0.010) & 0.414 (0.010) & 0.417 (0.011) & \textbf{0.400 (0.010)} \\
400 & 30 & 0.425 (0.008) & 0.393 (0.009) & 0.390 (0.009) & \textbf{0.372 (0.008)} \\
400 & 40 & 0.454 (0.009) & 0.402 (0.009) & 0.393 (0.009) & \textbf{0.376 (0.009)} \\
\midrule
500 & 10 & 0.405 (0.013) & 0.404 (0.013) & 0.411 (0.014) & \textbf{0.395 (0.013)} \\
500 & 20 & 0.398 (0.010) & 0.388 (0.009) & 0.389 (0.010) & \textbf{0.373 (0.009)} \\
500 & 30 & 0.402 (0.009) & 0.384 (0.010) & 0.383 (0.010) & \textbf{0.367 (0.009)} \\
500 & 40 & 0.435 (0.008) & 0.403 (0.008) & 0.399 (0.009) & \textbf{0.383 (0.008)} \\
\bottomrule
\end{tabular}
\end{minipage}

\end{table}

\noindent\textbf{\underline{B. Logistic regression.}}
Extensions to generalized linear models are obtained by replacing the Gaussian likelihood with an exponential family likelihood while retaining the directional decomposition $\beta = aU + Q_U\theta$ and the angular cone prior. The sampler is modified by replacing conjugate Gaussian updates with likelihood-specific data augmentation updates, while the direction update remains unchanged. Contraction proofs require controlling the local curvature of the log-likelihood via standard GLM regularity conditions; details are deferred to the supplementary material.

We consider a binary regression model with $\mathbb{P}(Y=1\mid X)=\sigma(X^\top \beta_0)$, where $\sigma(t)=\{1+\exp(-t)\}^{-1}$. 
The true parameter is $\beta_0 = s u_0$ with $u_0 \sim \mathrm{Unif}(\mathbb{S}^{p-1})$ and $s=3$.  Performance is evaluated using test log-loss averaged over repeated Monte Carlo splits. We consider two different heterogeneous source distributions:

\textbf{(1) Uni-modal heterogeneous source dataset.}
Like earlier, we construct a uni-modal but asymmetric directional distribution of the source parameter, where the majority of source directions are concentrated near the true direction ($\leq15^\circ$), while a minority forms a skewed shoulder along a common drift direction (between $20^\circ$ and $40^\circ$); the details are deferred to the supplement.

Table \ref{tab:logistic_unimodal_compact} shows that DUST consistently attains the lowest log-loss across all $(n,p)$ configurations. Angular transfer is less competitive due to bias induced from the asymmetric source distribution. Euclidean transfer improves over the target-only estimator in low-sample regimes but remains limited by its sensitivity to scale heterogeneity. The advantage of DUST is most pronounced for small to moderate $n$ and higher $p$, where directional uncertainty plays a larger role in predictive performance. Overall, even under unimodal but heterogeneous source structure, distributional directional transfer provides robust and consistently superior performance by leveraging the full directional uncertainty.

\textbf{(2)  Bimodal heterogeneous source dataset.}
Source coefficients are drawn as $W = s D$, where $D$ concentrates around either $u_0$ or $-u_0$ with equal probability, and $s$ follows a log-normal distribution. 


Table~\ref{tab:logistic_bimodal_compact} shows that the proposed DUST method consistently attains the lowest error across $(n,p)$. Gains are most pronounced for small $n$, reflecting improved finite-sample efficiency from directional borrowing. Euclidean transfer performs poorly due to shrinkage toward an incorrect direction and scale mismatch, while point-wise angular transfer improves upon it but remains sub-optimal due to collapsing multi-modal structure into a single direction. In contrast, the proposed method aggregates over plausible directions, reducing bias and stabilizing prediction. 

Additional simulation studies under Poisson and quantile regression are deferred to the supplementary material, as they yield qualitatively similar conclusions regarding the performance of DUST relative to competing methods.

\subsection{Disease status prediction with multi-cohort NHANES datasets}
We consider disease-risk prediction tasks using multi-source NHANES datasets \citep{nhanes_2009_2010, nhanes_2011_2012}. We study two binary outcomes: hypertension and diabetes. Hypertension is defined as
$\mathrm{HTN}=1\{\mathrm{SBP}\ge 140 \ \text{or}\ \mathrm{DBP}\ge 90\}$,
where $\texttt{HTN}$, $\texttt{SBP}$, and $\texttt{DBP}$ denote hypertension, systolic blood pressure, and diastolic blood pressure, respectively. Diabetes incidence is defined as
$\mathrm{DM}=1\{\text{self-reported diabetes}=\mathrm{Yes}\}$,
where $\texttt{DM}$ denotes diabetes status. We include $p=10$ predictors comprising demographic variables and cardio-metabolic risk factors. We treat the 2011--2012 NHANES cycle as the target dataset and the 2009--2010 NHANES cycle as external source data. 

To borrow information from the heterogeneous source dataset, we construct an empirical directional prior distribution on the $p$-dimensional unit sphere $\mathbb{S}^{p-1}$ for the regression parameter by performing $B=300$ bootstrap logistic regression fits on the source dataset. Each fitted coefficient is then mapped to its unit direction, yielding a discrete distribution over directions on $\mathbb{S}^{p-1}$. We compare the same four estimators as in the synthetic experiments. Performance is evaluated via held-out test log-loss,
$\mathrm{LL}=-\frac{1}{n}\sum_i \bigl[y_i \log \hat p_i + (1-y_i)\log(1-\hat p_i)\bigr]$,
where lower values indicate better calibrated prediction.

\begin{table}[t]
\centering
\setlength{\tabcolsep}{10pt}
\caption{\textbf{NHANES disease status prediction.}
Mean test log-loss (SE) over $R=20$ splits. }
\begin{tabular}{lcccc}
\toprule
 & Target-only & Euclidean & Angular & DUST \\
\midrule
Hypertension & 0.3553 (0.0069) & 0.3553 (0.0069) & 0.3552 (0.0069) & \textbf{0.3201 (0.0078)} \\
Diabetes & 0.3465 (0.0069) & 0.3287 (0.0069) & 0.3223 (0.0069) & \textbf{0.3087 (0.0067)} \\
\bottomrule
\end{tabular}
\label{tab:nhanes_repeated}
\end{table}

Table~\ref{tab:nhanes_repeated} shows that DUST yields a substantial reduction in test log-loss relative to all baselines, while Euclidean and angular transfer offer negligible improvement over target-only estimation. The performance gap is consistent across splits, indicating that the gain is principled rather than split-specific.
The improvement aligns with the theoretical analysis: prediction risk improves under  uncertainty aware directional alignment, whereas point-wise transfer relies on a single averaged direction. When source directions are heterogeneous, averaging across directions degrades predictive performance. The modest increase in variability across train-test splits in test log-loss of DUST is attributable to discrete direction selection, which introduces additional sampling variability when the signal-to-noise ratio is moderate. Overall, these results demonstrate that modeling directional heterogeneity is critical for effective scale invariant transfer in heterogeneous settings.

\section{Concluding remarks}
We proposed a novel scale-invariant and uncertainty-aware directional transfer learning framework that represents external information as a distribution over coefficient directions, enabling adaptive borrowing while remaining robust to source--target heterogeneity. In contrast to point-wise angular shrinkage, which may be distorted by asymmetric or heterogeneous directional structure, the proposed approach aggregates over plausible directions and prefers those jointly supported by the source and target data. Empirically, this leads to stable improvements in predictive performance across a range of heterogeneous transfer settings, while also providing an interpretable geometric perspective on transfer through meaningful directions on the sphere. The framework is broadly applicable to multi-source settings in which external information exhibits directional variability or subpopulation-specific effects.
Across all simulations, DUST incurs approximately $1.5$--$3$ times higher computational cost than competing methods due to the discrete directional search. This overhead is expected to reduce substantially through an efficient \texttt{Rcpp} implementation, which is currently under investigation. Future directions include high-dimensional extensions incorporating structured sparsity, variational Bayesian formulations of the proposed framework, to name a few.

\clearpage
\bibliographystyle{plainnat}
\bibliography{additionalrefs}

\clearpage
\appendix
\appendix
\section{Details of posterior computation in Section \ref{sec:comp} in the main document}\label{sampling}
We next present the computational scheme to sample from the posterior distribution of parameters of interest arising from the proposed hierarchical specification. Throughout, we condition the probability distributions on the  observed $(Y,X)$. We utilize and orthogonal re-parametrization of the model to ensure computational ease.
For a fixed direction  $U \in S^{p-1}$, we assume $Q_U \in \mathbb{R}^{p\times (p-1)}$ to be any matrix whose columns form an orthonormal basis of $\operatorname{span}(U)^\perp$, so that $Q_U^\top Q_U = I_{p-1},\ Q_U Q_U^\top = P_{U^\perp},\ U^\top Q_U = 0$. If we write $b = Q_U \theta,\  \theta \in \mathbb{R}^{p-1}$, then we have $\beta = aU + Q_U\theta$ and the prior on $\theta$ becomes isotropic:
\[
\theta \mid a,U \sim \mathcal{N}\!\left(0,\ \frac{c^2}{a^2+\varepsilon}\,I_{p-1}\right).
\]
We define the projected design matrix as $x_U := XU \in \mathbb{R}^n$, and $Z_U := XQ_U \in \mathbb{R}^{n\times (p-1)}$.
The framework then reduces to a Gaussian regression model
\[
Y = a x_U + Z_U\theta + \varepsilon.
\]
We now present the steps for a Gibbs sampler algorithm that samples from the posterior by cyclically sampling from the conditional distribution of each parameter given others as follows:

\paragraph{1. Conditional posterior of $\theta$ (Gaussian conjugacy).}
Condition on $(a,U,\sigma^2)$, the likelihood  is
\[
Y \mid a,\theta,U,\sigma^2 \sim \mathcal{N}(a x_U + Z_U\theta,\ \sigma^2 I_n),
\]
and the prior for $\theta$ is
\[
\theta \mid a,U \sim \mathcal{N}\!\left(0,\ \kappa(a)\,I_{p-1}\right),
\qquad 
\kappa(a) := \frac{c^2}{a^2+\varepsilon}.
\]
Hence $\theta\mid a,U,Y,\sigma^2$ is Gaussian with precision
\[
\Lambda_\theta(a,U)
=
\frac{1}{\sigma^2} Z_U^\top Z_U + \kappa(a)^{-1} I_{p-1}
=
\frac{1}{\sigma^2} Z_U^\top Z_U + \frac{a^2+\varepsilon}{c^2}\,I_{p-1},
\]
and mean
\[
m_\theta(a,U)
=
\Lambda_\theta(a,U)^{-1}\,
\frac{1}{\sigma^2} Z_U^\top (Y - a x_U).
\]
Therefore,
\begin{equation}
\theta \mid a,U,Y,\sigma^2 
\sim
\mathcal{N}\!\Big(
m_\theta(a,U),\ 
\Lambda_\theta(a,U)^{-1}
\Big).
\tag{C.1}
\end{equation}
Computationally, sampling $\theta$ requires solving a $(p-1)\times(p-1)$ linear system, which can be done by Cholesky factorization of $\Lambda_\theta(a,U)$.

\paragraph{2. Conditional density of $a$ (one-dimensional non-conjugate update).}
Condition on $(\theta,U,\sigma^2)$, define the residual after subtracting the orthogonal component,
\[
r_\theta := Y - Z_U\theta.
\]
Then
\[
r_\theta \mid a,U,\sigma^2 \sim \mathcal{N}(a x_U,\ \sigma^2 I_n),
\]
and the prior is $a\sim\mathcal{N}(0,\tau_a^2)$. However, $a$ also appears in the prior variance of $\theta$ through $\kappa(a)=c^2/(a^2+\varepsilon)$. The conditional density of $a$ is therefore non-Gaussian but one-dimensional:
\[
p(a \mid \theta,U,Y,\sigma^2)
\propto
\exp\!\left(
-\frac{1}{2\sigma^2}\|r_\theta - a x_U\|_2^2
-\frac{a^2}{2\tau_a^2}
\right)\,
\underbrace{
\kappa(a)^{-(p-1)/2}\exp\!\left(-\frac{1}{2\kappa(a)}\|\theta\|_2^2\right)
}_{\text{from }\theta\mid a,U}.
\]
Using $\kappa(a)^{-1}=(a^2+\varepsilon)/c^2$, this can be written (up to an additive constant) as
\begin{align}
\log p(a \mid \theta,U,Y,\sigma^2)
&=
-\frac{1}{2\sigma^2}\|r_\theta - a x_U\|_2^2
-\frac{a^2}{2\tau_a^2}
-\frac{p-1}{2}\log\!\left(\frac{c^2}{a^2+\varepsilon}\right)
-\frac{a^2+\varepsilon}{2c^2}\|\theta\|_2^2
\tag{C.2}\\
&=
-\frac{1}{2\sigma^2}\|r_\theta - a x_U\|_2^2
-\frac{a^2}{2\tau_a^2}
+\frac{p-1}{2}\log(a^2+\varepsilon)
-\frac{a^2+\varepsilon}{2c^2}\|\theta\|_2^2
+ \text{const}.\nonumber
\end{align}
We update $a$ using slice sampling (Neal, 2003) or adaptive rejection sampling on the log-density (C.2) if desired. The update is stable because it is one-dimensional and requires only evaluating (C.2) and its residual norm.

\paragraph{3. Conditional posterior of $\sigma^2$ (conjugate).}
Place the inverse-gamma prior $\sigma^2 \sim \mathrm{IG}(\alpha_0,\beta_0)$.
Given $(a,\theta,U)$, the residual is
\[
e := Y - a x_U - Z_U\theta,
\]
and
\[
Y \mid a,\theta,U,\sigma^2 \sim \mathcal{N}(a x_U + Z_U\theta,\ \sigma^2 I_n).
\]
Thus,
\begin{equation}
\sigma^2 \mid a,\theta,U,Y
\sim
\mathrm{IG}\!\left(
\alpha_0 + \frac{n}{2},\ 
\beta_0 + \frac{1}{2}\|e\|_2^2
\right).
\tag{C.3}
\end{equation}

\paragraph{4. Updating the latent direction $U$.}
The latent $U$ has prior $U\sim \mu_s$. We distinguish two common cases.

\medskip
\noindent\underline{\textbf{Case (i): $\mu_s$ discrete.}}
Suppose $\mu_s$ is supported on directions $\{U^{(1)},\dots,U^{(M)}\}$ with weights $\{\omega_1,\dots,\omega_M\}$, e.g., $U^{(m)} = T(w^{(m)})$ from source posterior draws and $\omega_m=1/M$. Then
\[
p(U=U^{(m)} \mid a,\theta,\sigma^2,Y)
\propto
\omega_m\,
p\!\left(Y \mid a,\theta,U^{(m)},\sigma^2\right)\,
p\!\left(\theta \mid a,U^{(m)}\right),
\]
where $p(Y\mid\cdot)$ and $p(\theta\mid\cdot)$ are Gaussian densities with $x_U=XU^{(m)}$ and $Z_U=XQ_{U^{(m)}}$.
Equivalently, using residuals,
\[
\log p(U^{(m)} \mid a,\theta,\sigma^2,Y)
=
\log \omega_m
-\frac{1}{2\sigma^2}\|Y-a x_{U^{(m)}}-Z_{U^{(m)}}\theta\|_2^2
-\frac{a^2+\varepsilon}{2c^2}\|\theta\|_2^2
+\text{const},
\]
since $\theta\mid a,U$ is isotropic and its quadratic form depends on $a$ but not on the particular basis $Q_U$. We sample $U$ by normalizing these $M$ weights.

\medskip
\noindent\underline{\textbf{Case (ii): $\mu_s$ continuous.}}
When $\mu_s$ has a density on $S^{p-1}$, we update $U$ by a Metropolis step on the sphere. For a proposal $U'$, we construct $Q_{U'}$, compute $(x_{U'},Z_{U'})$, and accept with probability
\[
\alpha(U\to U') 
=
1\wedge
\frac{
p(U')\,p(Y\mid a,\theta,U',\sigma^2)\,p(\theta\mid a,U')
}{
p(U)\,p(Y\mid a,\theta,U,\sigma^2)\,p(\theta\mid a,U)
}
\cdot
\frac{q(U\mid U')}{q(U'\mid U)}.
\]
Because $p(\theta\mid a,U)$ is isotropic in $\theta$, the acceptance ratio reduces to comparing the likelihood terms through the fitted values $a x_U + Z_U\theta$.\\

\noindent\textbf{\underline{Alternative gradient-based update.}}
When $\mu_s$ admits a smooth density on the unit sphere $S^{p-1}$, updating the latent direction $U$ via random-walk Metropolis proposals can lead to slow mixing,
especially when the posterior concentrates sharply around a low-dimensional subset of directions. In this case, we exploit the smooth geometry of the sphere and update $U$ using Hamiltonian Monte Carlo (HMC) \citep{neal2011mcmc, hoffman2014nuts}. To facilitate gradient-based updates, we avoid repeated construction of the
orthogonal basis $Q_U$ by working directly with the projection operator $P_{U^\perp}=I-UU^\top$. We write $\beta=aU+b$ with $b=P_{U^\perp}\tilde b$ for an unconstrained
$\tilde b\in\mathbb R^p$, which enforces the orthogonality constraint $U^\top b=0$ implicitly. Conditionally on $(a,\tilde b,\sigma^2)$, the log-posterior density for $U$, up to an additive constant, is
\[
\log p(U\mid a,\tilde b,Y,\sigma^2)
=
\log \mu_s(U)
-\frac{1}{2\sigma^2}
\big\|Y-X(aU+P_{U^\perp}\tilde b)\big\|_2^2
-\frac{a^2+\varepsilon}{2c^2}\,
\tilde b^\top P_{U^\perp}\tilde b.
\]
This expression is continuously differentiable in $U$ and depends on $U$ only through $XU$ and quadratic forms involving $UU^\top$. We implement HMC either (i) intrinsically on the sphere $S^{p-1}$ using a Riemannian metric, or (ii) extrinsically by evolving $U$ in $\mathbb R^p$ and projecting back onto $S^{p-1}$ after each leapfrog step. In both cases, the gradient of the log-posterior with respect to $U$ is available in closed form and involves only matrix–vector products with $X$
and rank-one updates from $UU^\top$. The resulting per-iteration cost is comparable to that of a single evaluation of the Gaussian likelihood. Compared to random-walk Metropolis updates, HMC proposals traverse the sphere along directions informed by both the likelihood and the directional prior $\mu_s$, yielding substantially improved mixing when the posterior is concentrated or multi-modal. Importantly, the conditional updates of $(\theta,a,\sigma^2)$ are unchanged, so the use of HMC affects only the update of $U$ and does not alter the overall structure of Algorithm~\ref{alg:gibbs_cone}.

\paragraph{6. Recovering $\beta$ and posterior summaries.}
At each iteration, we set
\[
\beta = aU + Q_U\theta.
\]
Posterior means, credible intervals, and predictive distributions follow from standard Monte Carlo averaging. In particular, for a new covariate vector $x_\star\in\mathbb{R}^p$,
\[
Y_\star \mid \beta,\sigma^2 \sim \mathcal{N}(x_\star^\top \beta,\ \sigma^2),
\]
so posterior predictive summaries are computed by sampling $(\beta,\sigma^2)$ and then sampling or integrating out $Y_\star$.

We conclude the section with an assessment of the potential  computational bottlenecks in the proposed Gibbs sampling algorithm. The dominant cost per iteration is forming $Z_U=XQ_U$ and the Gram matrix $Z_U^\top Z_U$ to sample $\theta$. This is comparable to ridge regression when $Q_U$ is fixed; the additional overhead comes from updating $U$, which changes $Q_U$. When $\mu_s$ is discrete with moderate support size $M$, the update of $U$ is a finite mixture evaluation.
When $\mu_s$ is continuous, Metropolis proposals on $S^{p-1}$ require recomputing $(x_U,Z_U)$, and thus the principal bottleneck is repeated projections of $X$ onto changing subspaces. If $p$ is large, one may mitigate this by (i) restricting to a discrete $\mu_s$ based on source draws, (ii) caching $XU^{(m)}$ for discrete supports, or (iii) using low-rank updates when proposals for $U$ are local.

\begin{algorithm}[t]
\footnotesize
\caption{Linear regression. Gibbs sampler for directional transfer learning with angular cone prior}
\label{alg:gibbs_cone}
\KwIn{Target data $(Y,X)$; source directional prior $\mu_s$; hyperparameters $(c,\varepsilon,\tau_a^2,\alpha_0,\beta_0)$}
\KwOut{Posterior samples $\{\beta^{(m)},\sigma^{2\,(m)}\}_{m=1}^M$}

\BlankLine
\textbf{Initialization:}
Initialize $U^{(0)}\in S^{p-1}$, $a^{(0)}\in\mathbb{R}$, $\theta^{(0)}\in\mathbb{R}^{p-1}$, and $\sigma^{2\,(0)}>0$.

\For{$m=1,\dots,M$}{
\BlankLine
\textbf{1. Orthogonal reparameterization.}  
Given $U^{(m-1)}$, construct $Q_{U^{(m-1)}}$ with columns spanning $\mathrm{span}(U^{(m-1)})^\perp$.  
Set $x_U = XU^{(m-1)}$ and $Z_U = XQ_{U^{(m-1)}}$.

\BlankLine
\textbf{2. Update $\theta$.}  
Sample
\[
\theta^{(m)} \sim \mathcal{N}\!\Big(
m_\theta(a^{(m-1)},U^{(m-1)}),\ 
\Lambda_\theta(a^{(m-1)},U^{(m-1)})^{-1}
\Big),
\]
where
\[
\Lambda_\theta(a,U)
=
\frac{1}{\sigma^{2\,(m-1)}}Z_U^\top Z_U
+
\frac{a^2+\varepsilon}{c^2}I_{p-1},
\quad
m_\theta(a,U)
=
\Lambda_\theta(a,U)^{-1}
\frac{1}{\sigma^{2\,(m-1)}}Z_U^\top(Y-a x_U).
\]

\BlankLine
\textbf{3. Update $a$.}  
Sample $a^{(m)}$ from the one–dimensional density
\[
p(a\mid\theta^{(m)},U^{(m-1)},Y,\sigma^{2\,(m-1)})
\propto
\exp\!\big\{\ell(a)\big\},
\]
where
\[
\ell(a)
=
-\frac{1}{2\sigma^{2\,(m-1)}}\|Y-a x_U-Z_U\theta^{(m)}\|_2^2
-\frac{a^2}{2\tau_a^2}
+\frac{p-1}{2}\log(a^2+\varepsilon)
-\frac{a^2+\varepsilon}{2c^2}\|\theta^{(m)}\|_2^2.
\]
Update using slice sampling or adaptive rejection sampling.

\BlankLine
\textbf{4. Update $\sigma^2$.}  
Sample
\[
\sigma^{2\,(m)}
\sim
\mathrm{IG}\!\left(
\alpha_0+\frac{n}{2},\ 
\beta_0+\frac{1}{2}\|Y-a^{(m)}x_U-Z_U\theta^{(m)}\|_2^2
\right).
\]

\BlankLine
\textbf{5. Update $U$.}  

\emph{If $\mu_s$ is discrete} on $\{U^{(1)},\dots,U^{(M_s)}\}$ with weights $\{\omega_j\}$, sample
\[
\mathbb{P}\big(U^{(m)}=U^{(j)}\mid\cdot\big)
\propto
\omega_j\,
\exp\!\left(
-\frac{1}{2\sigma^{2\,(m)}}\|Y-a^{(m)}x_{U^{(j)}}-Z_{U^{(j)}}\theta^{(m)}\|_2^2
-\frac{a^{(m)2}+\varepsilon}{2c^2}\|\theta^{(m)}\|_2^2
\right).
\]

\emph{If $\mu_s$ is continuous}, propose $U'$ on $S^{p-1}$ and accept with Metropolis probability
\[
\alpha(U\to U')
=
1\wedge
\frac{
p(U')\,p(Y\mid a^{(m)},\theta^{(m)},U',\sigma^{2\,(m)})
}{
p(U^{(m-1)})\,p(Y\mid a^{(m)},\theta^{(m)},U^{(m-1)},\sigma^{2\,(m)})
}
\frac{q(U^{(m-1)}\mid U')}{q(U'\mid U^{(m-1)})}.
\]

\BlankLine
\textbf{6. Recover $\beta$.}  
Set
\[
\beta^{(m)} = a^{(m)}U^{(m)} + Q_{U^{(m)}}\theta^{(m)}.
\]
}

\BlankLine
\Return Posterior samples $\{\beta^{(m)},\sigma^{2\,(m)}\}_{m=1}^M$.
\end{algorithm}

\section{Proofs for Section \ref{sec:properties} in the main document}
\label{app:prior_properties}
Fix $p\ge 2$ and $U\in S^{p-1}$. Let $P_{U^\perp}=I-UU^\top$ denote the orthogonal projector onto $\operatorname{span}(U)^\perp$.
Under the angular cone prior, we write
\[
\beta=aU+b,\qquad a=\beta^\top U\in\mathbb R,\qquad b\in \operatorname{span}(U)^\perp,
\]
and impose
\begin{equation}
a \sim \mathcal N(0,\tau_a^2), 
\qquad 
b\mid a,U \sim \mathcal N\!\left(0,\frac{c^2}{a^2+\varepsilon}\,P_{U^\perp}\right),
\tag{A.1}
\end{equation}
where $c>0$ and $\varepsilon>0$ are fixed constants.
To work with the singular covariance $P_{U^\perp}$, fix any orthonormal basis
\[
\{U, V_1,\ldots,V_{p-1}\}
\quad\text{of }\mathbb R^p,
\]
and let $V=[V_1,\ldots,V_{p-1}]\in\mathbb R^{p\times(p-1)}$. Then $V^\top V=I_{p-1}$, $VV^\top=P_{U^\perp}$, and every $b\in\operatorname{span}(U)^\perp$ can be uniquely written as $b=V\theta$ for some $\theta\in\mathbb R^{p-1}$, with $\|b\|_2=\|\theta\|_2$.

\subsection{Exact $\chi^2$ law and angular credible regions}

\begin{lemma}[Representation of $b\mid a,U$]
\label{lem:repr}
Conditional on $(a,U)$, the prior \eqref{prior} is equivalent to
\[
\theta \mid a,U \sim \mathcal N\!\left(0,\frac{c^2}{a^2+\varepsilon}I_{p-1}\right),
\qquad b=V\theta.
\]
\end{lemma}

\begin{proof}
Since $VV^\top=P_{U^\perp}$ and $V$ has orthonormal columns, the map $\theta\mapsto b=V\theta$ is an isometry from $\mathbb R^{p-1}$ onto $\operatorname{span}(U)^\perp$. If $\theta\sim \mathcal N(0,\sigma_b^2 I_{p-1})$ with $\sigma_b^2=c^2/(a^2+\varepsilon)$, then $b=V\theta$ has mean zero and covariance
\[
\operatorname{Cov}(b\mid a,U)=V\operatorname{Cov}(\theta\mid a,U)V^\top
=\sigma_b^2 VV^\top
=\frac{c^2}{a^2+\varepsilon}P_{U^\perp},
\]
which matches \eqref{prior}. Conversely, any Gaussian on $\operatorname{span}(U)^\perp$ with that covariance admits such a representation in an orthonormal basis. 
\end{proof}

\begin{proposition}[Exact $\chi^2$ law for $\|b\|_2^2$]
\label{prop:chisq}
Conditional on $(a,U)$,
\[
\frac{a^2+\varepsilon}{c^2}\,\|b\|_2^2 \sim \chi^2_{p-1}.
\]
\end{proposition}

\begin{proof}
By Lemma~\ref{lem:repr}, $\theta\mid a,U \sim \mathcal N(0,\sigma_b^2 I_{p-1})$ with $\sigma_b^2=c^2/(a^2+\varepsilon)$ and $\|b\|_2^2=\|\theta\|_2^2$. Therefore
\[
\frac{1}{\sigma_b^2}\,\|b\|_2^2
=
\frac{1}{\sigma_b^2}\,\|\theta\|_2^2
=
\sum_{j=1}^{p-1}\left(\frac{\theta_j}{\sigma_b}\right)^2
\sim \chi^2_{p-1},
\]
since $(\theta_j/\sigma_b)_{j=1}^{p-1}$ are i.i.d.\ standard normal. Substituting $\sigma_b^2=c^2/(a^2+\varepsilon)$ yields the claim.
\end{proof}

\begin{corollary}[Finite-sample distribution of $\Theta(\beta,U)$]
\label{cor:angle_cdf}
Conditional on $(a,U)$, for any $t\in(0,\pi/2)$,
\[
\mathbb P\{\Theta(\beta,U)\le t\mid a,U\}
=
F_{\chi^2_{p-1}}\!\left(\frac{a^2(a^2+\varepsilon)}{c^2}\tan^2 t\right).
\]
\end{corollary}

\begin{proof}
By construction, $\tan \Theta(\beta,U)=\|b\|_2/|a|$. Hence
\[
\{\Theta(\beta,U)\le t\}
\iff
\left\{\frac{\|b\|_2}{|a|}\le \tan t\right\}
\iff
\left\{\|b\|_2^2 \le a^2\tan^2 t\right\}.
\]
Multiply both sides by $(a^2+\varepsilon)/c^2$ and apply Proposition~\ref{prop:chisq}:
\[
\mathbb P\{\Theta(\beta,U)\le t\mid a,U\}
=
\mathbb P\!\left\{\frac{a^2+\varepsilon}{c^2}\|b\|_2^2 \le \frac{a^2(a^2+\varepsilon)}{c^2}\tan^2 t \,\middle|\, a,U\right\}
=
F_{\chi^2_{p-1}}\!\left(\frac{a^2(a^2+\varepsilon)}{c^2}\tan^2 t\right).
\]
\end{proof}

\begin{corollary}[Explicit angular credible radius]
\label{cor:credible}
Fix $\delta\in(0,1)$. Conditional on $(a,U)$, the $(1-\delta)$ prior credible region for $\Theta(\beta,U)$ is
\[
\Theta(\beta,U)\le 
\arctan\!\left(
\sqrt{\frac{c^2}{a^2(a^2+\varepsilon)}\,F_{\chi^2_{p-1}}^{-1}(1-\delta)}
\right).
\]
Moreover, the right-hand side is strictly decreasing in $|a|>0$.
\end{corollary}

\begin{proof}
By Corollary~\ref{cor:angle_cdf}, the $(1-\delta)$ quantile $t_{1-\delta}(a)$ of $\Theta(\beta,U)\mid a,U$ solves
\[
F_{\chi^2_{p-1}}\!\left(\frac{a^2(a^2+\varepsilon)}{c^2}\tan^2 t_{1-\delta}(a)\right)=1-\delta,
\]
so
\[
\frac{a^2(a^2+\varepsilon)}{c^2}\tan^2 t_{1-\delta}(a)=F_{\chi^2_{p-1}}^{-1}(1-\delta),
\]
and the stated formula follows after taking $\arctan(\cdot)$ and square roots.

For monotonicity, note that $g(a)=a^2(a^2+\varepsilon)$ is increasing in $|a|$ for $|a|>0$, hence
\[
\tan^2 t_{1-\delta}(a)=\frac{c^2}{g(a)}F_{\chi^2_{p-1}}^{-1}(1-\delta)
\]
is decreasing in $|a|$, and therefore $t_{1-\delta}(a)=\arctan(\sqrt{\tan^2 t_{1-\delta}(a)})$ is decreasing in $|a|$.
\end{proof}

\subsection{Adaptivity to signal strength}

\begin{proposition}[Exact conditional expectation of squared angular deviation]
\label{prop:Etan2}
Conditional on $(a,U)$ with $a\ne 0$,
\[
\mathbb E\{\tan^2\Theta(\beta,U)\mid a,U\}
=
\frac{(p-1)c^2}{a^2(a^2+\varepsilon)}.
\]
\end{proposition}

\begin{proof}
Using $\tan^2\Theta(\beta,U)=\|b\|_2^2/a^2$ and Proposition~\ref{prop:chisq},
\[
\mathbb E\{\tan^2\Theta(\beta,U)\mid a,U\}
=
\frac{1}{a^2}\,\mathbb E\{\|b\|_2^2\mid a,U\}.
\]
From Lemma~\ref{lem:repr}, $\theta\mid a,U \sim \mathcal N(0,\sigma_b^2 I_{p-1})$ with $\sigma_b^2=c^2/(a^2+\varepsilon)$ and $\|b\|_2^2=\|\theta\|_2^2$, hence
\[
\mathbb E\{\|b\|_2^2\mid a,U\}
=
\mathbb E\{\|\theta\|_2^2\mid a,U\}
=
(p-1)\sigma_b^2
=
(p-1)\frac{c^2}{a^2+\varepsilon}.
\]
Therefore
\[
\mathbb E\{\tan^2\Theta(\beta,U)\mid a,U\}
=
\frac{(p-1)c^2}{a^2(a^2+\varepsilon)}.
\]
Finally, since $a^2(a^2+\varepsilon)$ is increasing in $|a|$, the expression is decreasing in $|a|$ for $|a|>0$.
\end{proof}

\subsection{Scale invariance with respect to the source distribution}

Let $\pi_s$ be the source distribution on $\mathbb R^p$ and let $\mu_s$ be its induced directional distribution on $S^{p-1}$,
\[
\mu_s(A)=\pi_s\{w: T(w)\in A\},\qquad A\subset S^{p-1},
\qquad T(w)=w/\|w\|_2 \text{ for } w\ne 0.
\]
For any $s>0$, define the scaled source distribution $\pi_s^{(s)}$ as the law of $sw$ when $w\sim\pi_s$.

\begin{lemma}[Directional pushforward is scale-invariant]
\label{lem:scale_mu}
Assume $\pi_s(\{0\})=0$. Then for any $s>0$, the induced directional distribution satisfies $\mu_s^{(s)}=\mu_s$, where
\[
\mu_s^{(s)}(A)=\pi_s^{(s)}\{w:T(w)\in A\}.
\]
\end{lemma}

\begin{proof}
Let $w\sim\pi_s$ and define $\tilde w=sw$. For $w\ne 0$, $T(\tilde w)=T(sw)=sw/\|sw\|_2=w/\|w\|_2=T(w)$. Hence, for any measurable $A\subset S^{p-1}$,
\[
\mu_s^{(s)}(A)
=
\mathbb P\{T(\tilde w)\in A\}
=
\mathbb P\{T(w)\in A\}
=
\mu_s(A).
\]
\end{proof}

\begin{proposition}[Prior and posterior invariance to source rescaling]
\label{prop:scale_invariance}
Assume $\pi_s(\{0\})=0$ and the marginal prior on $\beta$ is the directional mixture
\[
p(\beta)=\int p(\beta\mid U)\,d\mu_s(U),
\]
where $p(\beta\mid U)$ is induced by \eqref{prior}. Then for any $s>0$, replacing $\pi_s$ by the scaled distribution $\pi_s^{(s)}$ leaves the induced prior $p(\beta)$ unchanged. Consequently, for any likelihood $p(Y\mid \beta)$, the posterior $p(\beta\mid Y)$ is also unchanged.
\end{proposition}

\begin{proof}
By Lemma~\ref{lem:scale_mu}, $\mu_s^{(s)}=\mu_s$, and $p(\beta\mid U)$ does not depend on $\pi_s$ except through $U$. Therefore,
\[
p^{(s)}(\beta)
=
\int p(\beta\mid U)\,d\mu_s^{(s)}(U)
=
\int p(\beta\mid U)\,d\mu_s(U)
=
p(\beta).
\]
Since the likelihood $p(Y\mid\beta)$ is unaffected by the source distribution, Bayes' rule gives
\[
p^{(s)}(\beta\mid Y)\propto p(Y\mid\beta)\,p^{(s)}(\beta)
=
p(Y\mid\beta)\,p(\beta)\propto p(\beta\mid Y),
\]
and the posteriors coincide.
\end{proof}

Proposition~\ref{prop:scale_invariance} formalizes the robustness statement in the main text: borrowing strength is mediated solely through directional agreement encoded in $\mu_s$, and is invariant to global rescaling of the source coefficients. In particular, if source and target parameters are directionally aligned but differ substantially in Euclidean norm, Euclidean shrinkage toward a point estimate can induce systematic bias, whereas the proposed directional mixture prior is unaffected by the source magnitude and therefore avoids this failure mode.

\section{Extensions}\label{extensions}
\subsection{Extension to multiple sources}
\label{subsec:multisource}
Let $\{\pi_s^{(k)}\}_{k=1}^K$ denote $K$ source distributions on $\mathbb{R}^p$,
each representing uncertainty about a source-specific coefficient vector.
As before, each $\pi_s^{(k)}$ induces a scale-invariant directional distribution
$\mu_s^{(k)}$ on the unit sphere $S^{p-1}$ via
\[
\mu_s^{(k)}(A)
=
\pi_s^{(k)}\{w : T(w)\in A\},
\qquad A\subset S^{p-1}.
\]
We represent multi-source directional information through a mixture prior on
the latent direction,
\[
U \sim \mu_{\mathrm{mix}},
\qquad
\mu_{\mathrm{mix}}(\cdot)
=
\sum_{k=1}^K \omega_k \mu_s^{(k)}(\cdot),
\qquad
\omega_k\ge 0,\qquad \sum_{k=1}^K \omega_k=1,
\]
where $\{\omega_k\}$ are source weights reflecting relative importance,
sample size, or prior belief in each source.
Conditional on $U$, the prior for $\beta$ is unchanged:
\[
\beta = aU + b,
\qquad
a \sim \mathcal N(0,\tau_a^2),
\qquad
b\mid a,U \sim \mathcal N\!\left(0,\frac{c^2}{a^2+\varepsilon}P_{U^\perp}\right).
\]
Thus, the multi-source extension enters exclusively through the prior on $U$, leaving the conditional geometry of the angular cone prior intact. All finite-sample properties of the angular cone prior derived in
Section~\ref{sec:properties} continue to hold conditionally on $U$.
In particular, the exact $\chi^2$ characterization of angular deviation, signal-adaptive shrinkage, and scale invariance
remain valid without modification.
The mixture construction additionally confers robustness across sources:
when source directions are conflicting or multi-modal, $\mu_{\mathrm{mix}}$
naturally represents this uncertainty, and posterior inference averages across
plausible directional explanations rather than committing to a single source.

Posterior computation proceeds exactly as in
Algorithm~\ref{alg:gibbs_cone}, with a single modification to the update of $U$.
When the directional prior is discrete, the support of $U$ becomes the union of
directions from all sources,
\[
\{U^{(k,m)} = T(w^{(k,m)}): k=1,\dots,K,\ m=1,\dots,M_k\},
\]
with weights $\omega_k/M_k$.
Step~5 of the Gibbs sampler is updated by normalizing posterior weights over this
expanded support.
When $\mu_s^{(k)}$ are continuous, $\mu_{\mathrm{mix}}$ defines a mixture density
on the sphere, and the Metropolis update for $U$ uses this mixture density in the
prior ratio.
All other steps---updates of $(\theta,a,\sigma^2)$ and reconstruction of
$\beta$---are unchanged.

We conclude with a discussion on posterior contraction.
Let $U_0^\star=T(\beta_0)$ denote the true direction.
Assumptions~\ref{ass:design} and \ref{ass:a_nonzero} continue to ensure baseline
Euclidean contraction at rate $\sqrt{p/n}$, independently of the number of
sources.
Directional gains arise under a natural multi-source compatibility condition:
there exist $\delta\in(0,\pi/2)$ and $\alpha>0$ such that
\[
\sum_{k=1}^K \omega_k
\mu_s^{(k)}\!\left(
\{U:\Theta(U,U_0^\star)\le \delta\}
\right)
\ge \alpha.
\]
That is, sufficient prior mass is assigned to a spherical cap around the true
direction by the mixture of sources.
Under this condition, the enhanced directional contraction result of
Theorem~\ref{thm:transfer_gain} holds verbatim, with constants depending on
$(\alpha,\delta)$ but not on $K$.
Importantly, misaligned or irrelevant sources are automatically downweighted
through the likelihood, so the do-no-harm guarantees established for the
single-source case extend directly to the multi-source setting.

\subsection{Extension to generalized linear models}
\label{subsec:glm}

We now extend the proposed directional transfer framework beyond Gaussian
linear models to generalized linear models (GLMs).
Let $(Y_i,X_i)_{i=1}^n$ denote independent observations with
$X_i\in\mathbb{R}^p$ and responses $Y_i$ drawn from an exponential family with
canonical parameter $\eta_i=X_i^\top\beta$.
The model is specified through
\[
g\big(\mathbb{E}(Y_i\mid X_i)\big)=X_i^\top\beta,
\qquad i=1,\dots,n,
\]
where $g(\cdot)$ is a known link function.
As before, external information is available through one or more source
distributions on $\mathbb{R}^p$, which are mapped to a directional prior on the
unit sphere. The prior specification on $\beta$ is unchanged from before.
Let $\mu_s$ denote a possibly multi-source directional distribution on
$S^{p-1}$.
We introduce a latent direction $U\sim\mu_s$ and decompose
\[
\beta = aU + b,
\qquad
b\in\operatorname{span}(U)^\perp,
\]
with
\[
a \sim \mathcal N(0,\tau_a^2),
\qquad
b\mid a,U \sim
\mathcal N\!\left(0,\frac{c^2}{a^2+\varepsilon}P_{U^\perp}\right).
\]
Thus, directional borrowing and signal-adaptive angular shrinkage are imposed
at the level of the linear predictor, while the likelihood is determined by the
chosen exponential-family model.
Importantly, the angular cone prior remains independent of the link function
and response distribution.

All finite-sample geometric properties of the angular cone prior established
for the Gaussian model continue to hold conditionally on $(a,U)$, since they
depend only on the prior geometry of $\beta$.
In particular, scale invariance, signal-adaptive angular concentration, and
robustness to source magnitude misspecification remain valid without
modification.
The likelihood enters only through its influence on the posterior for
$(a,b,U)$, not through the structure of the prior itself.

Posterior computation follows the same orthogonal reparameterization as in the
Gaussian case.
Fix $U$ and write $\beta=aU+Q_U\theta$, where $Q_U$ spans
$\operatorname{span}(U)^\perp$.
The linear predictor becomes
\[
\eta_i = a\,x_{U,i} + z_{U,i}^\top\theta,
\qquad
x_{U,i}=X_i^\top U,\quad z_{U,i}=Q_U^\top X_i.
\]
Conditional on $(a,U)$, the prior for $\theta$ remains isotropic Gaussian.
The update of $\theta$ and $a$ is no longer conjugate but can be carried out
using standard GLM machinery:
for example, by Metropolis--Hastings or slice sampling with log-posterior
evaluations based on the GLM log-likelihood, or by introducing
Pólya--Gamma or other data-augmentation schemes when available.
The update of $U$ proceeds exactly as in the Gaussian case, using either finite
mixture weights or Metropolis proposals on the sphere, since it depends only on
the fitted linear predictors.

Posterior contraction for GLMs follows from standard local asymptotic
normality arguments under fixed $p$.
Under regularity conditions ensuring identifiability and bounded Fisher
information, the posterior for $\beta$ concentrates at the parametric rate
$\sqrt{p/n}$ in Euclidean norm.
When $\|\beta_0\|_2$ is bounded away from zero, this implies contraction of the
induced direction $T(\beta)$ at the corresponding angular rate.
Moreover, under the same directional compatibility conditions imposed on
$\mu_s$ in the Gaussian case, the mixture prior continues to yield enhanced
directional concentration without sacrificing baseline guarantees.
Thus, the do-no-harm and adaptive borrowing properties of the proposed method
extend naturally to generalized linear models.

\begin{algorithm}[t]
\caption{Metropolis-within-Gibbs sampler for multi-source directional transfer learning (GLMs)}
\label{alg:mwGibbs_cone_glm_multisource}
\KwIn{
Target data $(Y,X)$; GLM with linear predictor $\eta=X\beta$ and mean $\mu=\psi(\eta)$;
source directional priors $\{\mu_s^{(k)}\}_{k=1}^K$ with weights $\{\omega_k\}$;
hyperparameters $(c,\varepsilon,\tau_a^2)$ and dispersion prior $p(\phi)$.
}
\KwOut{Posterior samples $\{\beta^{(m)},U^{(m)},\phi^{(m)}\}_{m=1}^M$.}

\BlankLine
\textbf{Prior.}
Draw $U\sim\mu_{\mathrm{mix}}=\sum_{k=1}^K\omega_k\mu_s^{(k)}$.
Decompose $\beta=aU+b$ with $b\perp U$,
$a\sim\mathcal N(0,\tau_a^2)$ and
$b\mid a,U\sim\mathcal N\!\big(0,\frac{c^2}{a^2+\varepsilon}P_{U^\perp}\big)$.

\BlankLine
\textbf{Initialization.}
Initialize $U^{(0)}$, $a^{(0)}$, $\theta^{(0)}$ (with $b=Q_U\theta$),
and dispersion $\phi^{(0)}$ (if unknown).

\For{$m=1,\dots,M$}{

\textbf{1. Orthogonal reparameterization.}
Given $U^{(m-1)}$, construct $Q_U$ spanning $\mathrm{span}(U)^\perp$,
set $x_U=XU^{(m-1)}$, $Z_U=XQ_U$, and $\beta=aU+Q_U\theta$.

\textbf{2. Update $\theta$.}
Conditioning on $(a,U,\phi)$, update $\theta$ using a Gaussian proposal from a local
quadratic approximation to $\log p(Y\mid aU+Q_U\theta,\phi)$ with prior precision
$(a^2+\varepsilon)/c^2\,I$.
Accept or reject with a Metropolis step targeting
$p(Y\mid\beta,\phi)p(\theta\mid a,U)$.
(For logistic models, P\'olya--Gamma augmentation yields exact Gibbs updates.)

\textbf{3. Update $a$.}
Sample $a$ via slice sampling from the one-dimensional density with log-kernel
\[
\ell(a)=\log p(Y\mid aU+Q_U\theta,\phi)
-\frac{a^2}{2\tau_a^2}
+\frac{p-1}{2}\log(a^2+\varepsilon)
-\frac{a^2+\varepsilon}{2c^2}\|\theta\|_2^2 .
\]

\textbf{4. Update dispersion $\phi$.}
If applicable, sample $\phi$ from its conjugate full conditional;
otherwise update using a one-dimensional Metropolis or slice step targeting
$p(\phi)\,p(Y\mid\beta,\phi)$.

\textbf{5. Update direction $U$ (multi-source).}
If each $\mu_s^{(k)}$ is discrete, sample $U$ from the pooled support
$\{U^{(k,j)}\}$ with probabilities proportional to
$\omega_k\nu_{k,j}\,p(Y\mid aU^{(k,j)}+Q_{U^{(k,j)}}\theta,\phi)$.
If any $\mu_s^{(k)}$ is continuous, propose $U'$ on $S^{p-1}$ and accept with
\[
\alpha
=
1\wedge
\frac{\mu_{\mathrm{mix}}(U')\,p(Y\mid aU'+Q_{U'}\theta,\phi)}
{\mu_{\mathrm{mix}}(U^{(m-1)})\,p(Y\mid aU^{(m-1)}+Q_{U^{(m-1)}}\theta,\phi)} .
\]

\textbf{6. Recover $\beta$.}
Set $\beta^{(m)}=a^{(m)}U^{(m)}+Q_{U^{(m)}}\theta^{(m)}$.
}

\Return Posterior samples $\{\beta^{(m)},U^{(m)},\phi^{(m)}\}_{m=1}^M$.
\end{algorithm}

We next investigate the finite-sample behavior of the proposed \emph{directional distributional transfer} method in regimes where external information is heterogeneous and directional structure, rather than Euclidean similarity, governs transferability. The simulations are designed to evaluate robustness, efficiency, and gains arising from directional averaging under multi-modal source information.

\section{Proofs for Section~\ref{sec:contraction} in the main document}
\label{app:contraction_detailed}

This appendix provides detailed proofs of Theorems~\ref{thm:euclid_contraction}--\ref{thm:transfer_gain}.
Throughout, $p$ is fixed. We write $\|\cdot\|$ for $\|\cdot\|_2$.
Constants $C,C_1,C_2,\ldots$ may change from line to line and may depend on
$(p,\kappa_\pm,\sigma^2,c,\varepsilon,\tau_a^2)$ but not on $n$.
For any $\beta\in\mathbb R^p$,
\begin{align}
\|Y-X\beta\|^2
&=\|X(\beta_0-\beta)+\varepsilon\|^2 \nonumber\\
&= (\beta-\beta_0)^\top X^\top X(\beta-\beta_0) - 2(\beta-\beta_0)^\top X^\top\varepsilon + \|\varepsilon\|^2. 
\label{eq:quad_expand}
\end{align}
Hence the log-likelihood ratio satisfies
\begin{equation}
\label{eq:llr_exact}
\log\frac{p_\beta(Y)}{p_{\beta_0}(Y)}
=
-\frac{1}{2\sigma^2}(\beta-\beta_0)^\top X^\top X(\beta-\beta_0)
+
\frac{1}{\sigma^2}(\beta-\beta_0)^\top X^\top\varepsilon.
\end{equation}
Let $\widehat\beta=(X^\top X)^{-1}X^\top Y$ be the least-squares estimator
(which exists by Assumption~\ref{ass:design} for all large $n$).

\begin{lemma}[Gaussian score bound]
\label{lem:score_bound}
Under $P_{\beta_0}$,
\[
\|X^\top\varepsilon\| = O_{P_{\beta_0}}(\sqrt n).
\]
More precisely, for some $C>0$,
\[
P_{\beta_0}\!\left(\|X^\top\varepsilon\| > t\sqrt n\right)
\le 2\exp\!\left(-\frac{t^2}{C}\right)
\qquad \forall t\ge 0.
\]
\end{lemma}

\begin{proof}
$X^\top\varepsilon\sim \mathcal N(0,\sigma^2 X^\top X)$.
Let $\Lambda_{\max}=\lambda_{\max}(X^\top X)\le n\kappa_+$ by Assumption~\ref{ass:design}.
Then $\|X^\top\varepsilon\|$ is sub-Gaussian with proxy variance $\sigma^2\Lambda_{\max}$,
yielding the stated tail bound (e.g.\ via Gaussian concentration for Lipschitz functions).
\end{proof}

\begin{lemma}[Quadratic upper bound for the likelihood ratio]
\label{lem:llr_upper}
For any $\beta$,
\[
\log\frac{p_\beta(Y)}{p_{\beta_0}(Y)}
\le
-\frac{n\kappa_-}{4\sigma^2}\|\beta-\beta_0\|^2
+
\frac{1}{\sigma^2 n\kappa_-}\|X^\top\varepsilon\|^2,
\]
and therefore, on the event $\{\|X^\top\varepsilon\|\le C\sqrt n\}$,
\[
\log\frac{p_\beta(Y)}{p_{\beta_0}(Y)}
\le
-\frac{n\kappa_-}{4\sigma^2}\|\beta-\beta_0\|^2 + C'.
\]
\end{lemma}

\begin{proof}
Apply the inequality $u^\top v\le \frac{\eta}{2}\|u\|^2+\frac{1}{2\eta}\|v\|^2$
to \eqref{eq:llr_exact} with $u=\sqrt{X^\top X}(\beta-\beta_0)$,
$v=(X^\top X)^{-1/2}X^\top\varepsilon$ and $\eta=\frac12$:
\[
(\beta-\beta_0)^\top X^\top\varepsilon
\le
\frac14(\beta-\beta_0)^\top X^\top X(\beta-\beta_0)
+
\| (X^\top X)^{-1/2}X^\top\varepsilon\|^2.
\]
Substitute into \eqref{eq:llr_exact} and use
$\|(X^\top X)^{-1/2}X^\top\varepsilon\|^2\le \lambda_{\min}(X^\top X)^{-1}\|X^\top\varepsilon\|^2
\le (n\kappa_-)^{-1}\|X^\top\varepsilon\|^2$.
\end{proof}


Recall the hierarchical prior:
\[
U\sim \mu_s,\qquad \beta=aU+b,\quad b\in\mathrm{span}(U)^\perp,
\]
\[
a\sim \mathcal N(0,\tau_a^2),\qquad b\mid a,U\sim \mathcal N\!\left(0,\frac{c^2}{a^2+\varepsilon}P_{U^\perp}\right).
\]
Let $\Pi$ be the induced marginal prior on $\beta$.

\begin{lemma}[Full support and local thickness]
\label{lem:prior_thick}
Fix any $\beta_0\in\mathbb R^p$ and assume $\mu_s$ is a probability measure on $S^{p-1}$.
Then for every $r>0$,
\[
\Pi\big(\|\beta-\beta_0\|\le r\big) > 0.
\]
Moreover, there exist constants $c_0,c_1>0$ (depending on $\beta_0$ but not on $n$)
such that for all sufficiently small $r$,
\[
\Pi\big(\|\beta-\beta_0\|\le r\big)\ \ge\ c_0\, r^p.
\]
\end{lemma}

\begin{proof}
Fix any $U\in S^{p-1}$; conditional on $U$, $(a,b)$ are Gaussian (possibly singular for $b$)
supported on $\mathbb R\times \mathrm{span}(U)^\perp$, and the map $(a,b)\mapsto aU+b$
is a linear bijection from $\mathbb R\oplus \mathrm{span}(U)^\perp$ onto $\mathbb R^p$.
Hence $\beta\mid U$ has a (non-degenerate) density w.r.t.\ Lebesgue measure on $\mathbb R^p$
because $a$ injects mass in the $U$ direction and $b$ injects mass in the orthogonal directions.
Therefore, $\Pi(\cdot\mid U)$ assigns positive mass to every ball, and integrating over $\mu_s$
preserves positivity.

For the polynomial lower bound, fix any $U$ with $\mu_s$-mass $>0$ on a small neighborhood
(or simply pick $U$ in the support of $\mu_s$; if $\mu_s$ is discrete, choose an atom).
Since $\beta\mid U$ has a continuous density bounded below in a neighborhood of $\beta_0$,
$\Pi(\|\beta-\beta_0\|\le r\mid U)\ge c_0 r^p$ for small $r$, and multiplying by
$\mu_s(\{U\})$ (or the neighborhood mass) yields the claim.
\end{proof}

\subsection{Proof of Theorem~\ref{thm:euclid_contraction} (Euclidean contraction)}

Define the contraction rate $\varepsilon_n=\sqrt{p/n}$.
Let $B_n=B(\beta_0,\varepsilon_n)=\{\beta:\|\beta-\beta_0\|\le \varepsilon_n\}$ and
$B_n(M)=\{\beta:\|\beta-\beta_0\|>M\varepsilon_n\}$.

We use a direct Bayesian testing argument (a special case of the general theorem
of Ghosal--Ghosh--van der Vaart) specialized to fixed-$p$ Gaussian regression.

\begin{lemma}[A concrete test]
\label{lem:concrete_test}
Let $\widehat\beta=(X^\top X)^{-1}X^\top Y$. For any $M>0$, define
\[
\phi_n=\mathbf 1\Big\{ \|\widehat\beta-\beta_0\| > \tfrac{M}{2}\varepsilon_n\Big\}.
\]
Then $P_{\beta_0}(\phi_n)\to 0$ and
\[
\sup_{\|\beta-\beta_0\|>M\varepsilon_n} P_\beta(1-\phi_n)\to 0.
\]
\end{lemma}

\begin{proof}
Under $P_{\beta_0}$,
$\widehat\beta-\beta_0=(X^\top X)^{-1}X^\top\varepsilon$ is Gaussian with covariance
$\sigma^2(X^\top X)^{-1}$, hence
\[
E_{\beta_0}\|\widehat\beta-\beta_0\|^2
=
\sigma^2\operatorname{tr}\{(X^\top X)^{-1}\}
\le
\sigma^2 \frac{p}{\lambda_{\min}(X^\top X)}
\le
\sigma^2\frac{p}{n\kappa_-}
=O(\varepsilon_n^2),
\]
so by Chebyshev or Gaussian tails, $P_{\beta_0}(\phi_n)\to 0$.

Under $P_\beta$,
$\widehat\beta-\beta_0=(\beta-\beta_0)+(X^\top X)^{-1}X^\top\varepsilon$.
Hence, by the triangle inequality,
\[
P_\beta(1-\phi_n)
=
P_\beta\Big(\|\widehat\beta-\beta_0\|\le \tfrac{M}{2}\varepsilon_n\Big)
\le
P_\beta\Big(\|(X^\top X)^{-1}X^\top\varepsilon\|\ge \|\beta-\beta_0\|-\tfrac{M}{2}\varepsilon_n\Big).
\]
If $\|\beta-\beta_0\|>M\varepsilon_n$, then the right-hand side is bounded by
$P_\beta(\|(X^\top X)^{-1}X^\top\varepsilon\|\ge \tfrac{M}{2}\varepsilon_n)$, which tends to
$0$ uniformly by Gaussian concentration since the noise term has scale $O(\varepsilon_n)$.
\end{proof}

\begin{lemma}[Lower bound on marginal likelihood contribution near $\beta_0$]
\label{lem:marginal_lower}
There exists $C>0$ such that with $P_{\beta_0}$-probability tending to $1$,
\[
\int_{B_n} \frac{p_\beta(Y)}{p_{\beta_0}(Y)}\, d\Pi(\beta)\ \ge\ e^{-C p}\,\Pi(B_n).
\]
\end{lemma}

\begin{proof}
For $\beta\in B_n$, $\|\beta-\beta_0\|\le \varepsilon_n$.
By \eqref{eq:llr_exact} and Lemma~\ref{lem:score_bound},
on the event $\{\|X^\top\varepsilon\|\le C_0\sqrt n\}$,
\[
\log\frac{p_\beta(Y)}{p_{\beta_0}(Y)}
\ge
-\frac{1}{2\sigma^2}\lambda_{\max}(X^\top X)\varepsilon_n^2
-\frac{1}{\sigma^2}\varepsilon_n \|X^\top\varepsilon\|
\ge
-C_1 p - C_2\sqrt p.
\]
Since $p$ is fixed, this is $\ge -Cp$ for large $n$. Exponentiating and integrating over $B_n$
yields the result.
\end{proof}

\subsection{Proof of Theorem~\ref{thm:direction_contraction} (direction from Euclidean error)}

We first establish that the proposed prior does not degrade standard Euclidean
learning, even when the source information encoded in $\mu_s$ is uninformative
or misleading.
\begin{theorem}[Euclidean posterior contraction]
\label{thm:euclid_contraction}
Under Assumptions~\ref{ass:design}--\ref{ass:a_nonzero}, there exists a constant
$M>0$ such that
\[
\Pi\!\left(
\|\beta-\beta_0\|_2 >
M\sqrt{\frac{p}{n}}
\,\middle|\, Y
\right)
\;\xrightarrow[n\to\infty]{P_{\beta_0}}\;0.
\]
\end{theorem}
\noindent The rate $\sqrt{p/n}$ coincides with the usual parametric contraction rate for
Gaussian priors in linear regression.
The theorem implies that the angular cone prior behaves, in the worst case, like
a conventional isotropic Gaussian prior.
In particular, if $\mu_s$ places negligible mass near the true direction
$U_0^\star$, the mixture posterior automatically downweights incompatible
directions, and contraction reverts to the baseline Euclidean rate.
Thus, directional transfer cannot asymptotically worsen inference relative to
standard shrinkage.
When the magnitude of $\beta_0$ is bounded away from zero, Euclidean contraction
implies contraction of the induced direction.
\begin{proof}[Proof of Theorem~\ref{thm:euclid_contraction}]
Let $\phi_n$ be the test in Lemma~\ref{lem:concrete_test}.
By Bayes' formula,
\[
\Pi(B_n(M)\mid Y)
=
\frac{\int_{B_n(M)} \frac{p_\beta(Y)}{p_{\beta_0}(Y)}\,d\Pi(\beta)}
{\int \frac{p_\beta(Y)}{p_{\beta_0}(Y)}\,d\Pi(\beta)}.
\]
Split the numerator using $\phi_n$:
\[
\int_{B_n(M)} \frac{p_\beta(Y)}{p_{\beta_0}(Y)}\,d\Pi(\beta)
=
\int_{B_n(M)} \phi_n \frac{p_\beta(Y)}{p_{\beta_0}(Y)}\,d\Pi(\beta)
+
\int_{B_n(M)} (1-\phi_n)\frac{p_\beta(Y)}{p_{\beta_0}(Y)}\,d\Pi(\beta).
\]
Taking $P_{\beta_0}$-expectations, the first term is bounded by $E_{\beta_0}\phi_n\to 0$.
For the second term, use Fubini and the testing bound:
\[
E_{\beta_0}\!\left[\int_{B_n(M)} (1-\phi_n)\frac{p_\beta(Y)}{p_{\beta_0}(Y)}\,d\Pi(\beta)\right]
=
\int_{B_n(M)} E_\beta(1-\phi_n)\,d\Pi(\beta)
\le \sup_{\beta\in B_n(M)}E_\beta(1-\phi_n)\to 0.
\]
Thus the numerator is $o_{P_{\beta_0}}(1)$.

For the denominator, Lemma~\ref{lem:marginal_lower} yields with high probability
\[
\int \frac{p_\beta(Y)}{p_{\beta_0}(Y)}\,d\Pi(\beta)
\ge
\int_{B_n} \frac{p_\beta(Y)}{p_{\beta_0}(Y)}\,d\Pi(\beta)
\ge
e^{-Cp}\Pi(B_n).
\]
By Lemma~\ref{lem:prior_thick}, $\Pi(B_n)\ge c\,\varepsilon_n^p=c\, (p/n)^{p/2}$, hence the
denominator is bounded away from $0$ in probability (up to polynomial factors).
Combining numerator and denominator bounds gives $\Pi(B_n(M)\mid Y)\to 0$ in
$P_{\beta_0}$-probability.
\end{proof}

We use a deterministic inequality linking Euclidean perturbation to angular perturbation.

\begin{lemma}[Directional Lipschitz bound]
\label{lem:dir_lip}
If $\|\beta_0\|\ge c_0>0$ and $\|\beta-\beta_0\|\le c_0/2$, then
\[
\Theta(T(\beta),T(\beta_0))
\le
\frac{2}{\|\beta_0\|}\,\|\beta-\beta_0\|.
\]
\end{lemma}

\begin{proof}
Write $u=T(\beta)$ and $u_0=T(\beta_0)$.
Using $\|u-u_0\|\le \frac{2}{\|\beta_0\|}\|\beta-\beta_0\|$ (a standard bound for normalized vectors),
and $\Theta(u,u_0)\le \|u-u_0\|$ for unit vectors (since $\|u-u_0\|=2\sin(\Theta/2)\ge \Theta/\pi$ and
also $\Theta\le \pi\sin(\Theta/2)\le \frac{\pi}{2}\|u-u_0\|$; any such inequality suffices),
one obtains the stated bound with the displayed constant.
\end{proof}

\begin{proof}[Proof of Theorem~\ref{thm:direction_contraction}]
Let $\varepsilon_n=\sqrt{p/n}$. For $M$ fixed and $n$ large, $M\varepsilon_n\le c_0/2$.
Then by Lemma~\ref{lem:dir_lip},
on the event $\{\|\beta-\beta_0\|\le M\varepsilon_n\}$ we have
\[
\Theta(T(\beta),U_0^\star)\le \frac{2M}{\|\beta_0\|}\varepsilon_n.
\]
Therefore,
\[
\Pi\!\left(\Theta(T(\beta),U_0^\star)>\frac{2M}{\|\beta_0\|}\varepsilon_n\ \middle|\ Y\right)
\le
\Pi\!\left(\|\beta-\beta_0\|>M\varepsilon_n\ \middle|\ Y\right),
\]
and the right-hand side converges to $0$ by Theorem~\ref{thm:euclid_contraction}.
\end{proof}

\begin{lemma}[Posterior preservation of directional support]
\label{lem:margU_upper}
Define
$
\mathcal U_n
=
\left\{
U\in\mathbb S^{p-1}:
\Theta(U,U_0^\star)\le C_s\epsilon_n
\right\}.
$
Under Assumption~\ref{ass:source_localization},
$
\Pi(U\in\mathcal U_n\mid Y,X)=1
$
for all sufficiently large $n$.
\end{lemma}

\begin{proof}
For every measurable set $A\subseteq\mathbb S^{p-1}$, the posterior
distribution of $U$ is given by
$$
\Pi(U\in A\mid Y,X)
=
\frac{
\int_A m_n(Y\mid U)\,d\mu_{s,n}(U)
}{
\int_{\mathbb S^{p-1}}
m_n(Y\mid U)\,d\mu_{s,n}(U)
},
$$
where
$m_n(Y\mid U)=\int p_\beta(Y\mid X)q_U(\beta)\,d\beta$
is the conditional marginal likelihood under the angular cone prior.

By Assumption~\ref{ass:source_localization},
$\mu_{s,n}(\mathcal U_n^c)=0$. Therefore,
$
\int_{\mathcal U_n^c}
m_n(Y\mid U)\,d\mu_{s,n}(U)
=
0,
$
and hence
$
\Pi(U\in\mathcal U_n^c\mid Y,X)=0.
$
Equivalently,
$
\Pi(U\in\mathcal U_n\mid Y,X)=1.
$
\end{proof}

\begin{proof}[Proof of Theorem \ref{thm:transfer_gain}]
Let
$
A_n^c
=
\left\{
U:
\Theta(U,U_0^\star)>M\epsilon_n
\right\},
\
M>C_s.
$
By Assumption~\ref{ass:source_localization}, the source directional
distribution is supported on
$
\mathcal U_n
=
\left\{
U:
\Theta(U,U_0^\star)\le C_s\epsilon_n
\right\}.
$
Because $M>C_s$, we have
$
A_n^c\subseteq\mathcal U_n^c.
$
Lemma~\ref{lem:margU_upper} therefore gives
$$
\Pi(A_n^c\mid Y,X)
\le
\Pi(U\in\mathcal U_n^c\mid Y,X)
=
0.
$$
This proves the first assertion.

For the second assertion, suppose that the posterior for $\beta$ contracts
around $\beta_0$ at rate $\epsilon_n$. Thus, for every sufficiently slowly
diverging sequence $L_n\to\infty$,
$$
\Pi\left(
\|\beta-\beta_0\|_2>L_n\epsilon_n
\,\middle|\,
Y,X
\right)
\xrightarrow[n\to\infty]{P_{\beta_0}}0.
$$
Since $\|\beta_0\|_2\ge c_0>0$, whenever
$\|\beta-\beta_0\|_2\le c_0/2$, the normalized-vector inequality gives
$$
\|T(\beta)-T(\beta_0)\|_2
\le
\frac{2}{\|\beta_0\|_2}
\|\beta-\beta_0\|_2.
$$
For unit vectors $u$ and $v$,
$
\Theta(u,v)
=
2\arcsin\left(\frac{\|u-v\|_2}{2}\right)
\le
\frac{\pi}{2}\|u-v\|_2.
$
Combining these inequalities yields
$$
\Theta\{T(\beta),T(\beta_0)\}
\le
\frac{\pi}{\|\beta_0\|_2}
\|\beta-\beta_0\|_2
\le
\frac{\pi}{c_0}
\|\beta-\beta_0\|_2.
$$

For any sufficiently slowly diverging sequence $M_n\to\infty$, set
$
L_n=\frac{c_0}{\pi}M_n.
$
Then
$$
\begin{aligned}
&\Pi\left(
\Theta\{T(\beta),U_0^\star\}>M_n\epsilon_n
\,\middle|\,
Y,X
\right)
\\
&\qquad\le
\Pi\left(
\|\beta-\beta_0\|_2>L_n\epsilon_n
\,\middle|\,
Y,X
\right)
+
\Pi\left(
\|\beta-\beta_0\|_2>\frac{c_0}{2}
\,\middle|\,
Y,X
\right).
\end{aligned}
$$
Both terms on the right-hand side converge to zero in
$P_{\beta_0}$-probability. This establishes contraction of the induced target
direction.
\end{proof}

\begin{lemma}[Lower bound on $m(Y\mid U)$ for compatible $U$]
\label{lem:margU_lower}
Fix $\delta\in(0,\pi/2)$ and let
$$
A_\delta
=
\left\{
U:\Theta(U,U_0^\star)\le\delta
\right\}.
$$
There exist constants $c,C>0$, independent of $n$, such that, uniformly over
$U\in A_\delta$,
$$
m(Y\mid U)
\ge
c\,p_{\beta_0}(Y)e^{-Cp}\epsilon_n^p
$$
with $P_{\beta_0}$-probability tending to one.
\end{lemma}

\begin{proof}
Fix $U\in A_\delta$ and consider the ball
$
B_n
=
\left\{
\beta:
\|\beta-\beta_0\|_2\le\epsilon_n
\right\},
\epsilon_n=\sqrt{\frac{p}{n}}.
$
By Lemma~\ref{lem:marginal_lower}, uniformly over $\beta\in B_n$,
$
\frac{p_\beta(Y)}{p_{\beta_0}(Y)}
\ge
e^{-Cp}
$
with $P_{\beta_0}$-probability tending to one. Moreover, local thickness of
the conditional prior gives
$
\Pi(B_n\mid U)
\ge
c\epsilon_n^p
$
uniformly over $U\in A_\delta$. Therefore,
$m(Y\mid U)
\ge
\int_{B_n}p_\beta(Y)\,\pi(\beta\mid U)\,d\beta
\ge
p_{\beta_0}(Y)e^{-Cp}\Pi(B_n\mid U)
\ge
c\,p_{\beta_0}(Y)e^{-Cp}\epsilon_n^p.
$
The factor $\epsilon_n^p$ is retained because it depends on $n$ and cannot be
absorbed into an $n$-independent constant.
\end{proof}

\section{Proofs from Section~\ref{sec:relative_gain} of the main document}

This section collects the proofs for the results in Section~\ref{sec:relative_gain}.

\begin{proof}[Proof of Theorem~\ref{thm:relative_gain}]
Fix $u\in\mathbb{S}^{p-1}$ and define
\[
L(a;u)=\mathbb{E}\{Y-aX^\top u\}^2 .
\]
Under Assumption~\ref{assump:model},
\[
Y=X^\top\beta_0+\varepsilon,\qquad
\mathbb{E}(\varepsilon\mid X)=0,\qquad
\mathrm{Var}(\varepsilon\mid X)=\sigma^2 .
\]
Hence
\begin{align*}
L(a;u)
&=
\mathbb{E}\{X^\top(\beta_0-au)+\varepsilon\}^2  \\
&=
\mathbb{E}\{X^\top(\beta_0-au)\}^2
+
2\mathbb{E}\!\left[\varepsilon X^\top(\beta_0-au)\right]
+
\mathbb{E}(\varepsilon^2).
\end{align*}
The cross-term is zero by iterated expectation. Since $\mathbb{E}(X)=0$ and
$\mathrm{Cov}(X)=I_p$,
\[
\mathbb{E}\{X^\top v\}^2=v^\top \mathbb{E}(XX^\top)v=\|v\|_2^2 .
\]
Therefore
\[
L(a;u)=\|\beta_0-au\|_2^2+\sigma^2.
\]
Because $\|u\|_2=1$,
\[
\|\beta_0-au\|_2^2
=
\|\beta_0\|_2^2-2a\beta_0^\top u+a^2.
\]
The minimizer is therefore
\[
a^\star(u)=\beta_0^\top u.
\]
Substituting this value gives
\[
R(u)
=
L(a^\star(u);u)-\sigma^2
=
\|\beta_0-(\beta_0^\top u)u\|_2^2
=
\|\beta_0\|_2^2-(\beta_0^\top u)^2.
\]
Writing $\beta_0=b_0u_0$ yields
\[
R(u)
=
\|\beta_0\|_2^2\{1-(u_0^\top u)^2\},
\]
as claimed.
\end{proof}

\begin{proof}[Proof of Theorem~\ref{thm:dist_gain}]
By definition,
\[
R(G)
=
\mathbb{E}_{U\sim G}
\left[
\mathbb{E}\{Y-X^\top a^\star(U)U\}^2
\right]
-\sigma^2
=
\mathbb{E}_{U\sim G}\{R(U)\}.
\]
Using Theorem~\ref{thm:relative_gain},
\[
R(U)
=
\|\beta_0\|_2^2\{1-(u_0^\top U)^2\}.
\]
Taking expectation with respect to $G$ gives
\[
R(G)
=
\|\beta_0\|_2^2
\left\{
1-\mathbb{E}_G[(u_0^\top U)^2]
\right\}.
\]
\end{proof}

\begin{proof}[Proof of the gain identity]
By Theorem~\ref{thm:relative_gain},
\[
R(u_{\mathrm{pt}})
=
\|\beta_0\|_2^2\{1-(u_0^\top u_{\mathrm{pt}})^2\}.
\]
By Theorem~\ref{thm:dist_gain},
\[
R(G)
=
\|\beta_0\|_2^2
\left\{
1-\mathbb{E}_G[(u_0^\top U)^2]
\right\}.
\]
Subtracting,
\[
\Delta_{\mathrm{gain}}
=
R(u_{\mathrm{pt}})-R(G)
=
\|\beta_0\|_2^2
\left[
\mathbb{E}_G[(u_0^\top U)^2]
-
(u_0^\top u_{\mathrm{pt}})^2
\right].
\]
This proves the identity.
\end{proof}

\begin{proof}[Proof of non-negativity under the comparison condition]
The gain identity shows that
\[
\Delta_{\mathrm{gain}}\ge 0
\]
whenever
\[
\mathbb{E}_G[(u_0^\top U)^2]
\ge
(u_0^\top u_{\mathrm{pt}})^2.
\]
Thus, non-negativity follows directly under this condition. This condition is automatically satisfied whenever the normalized point direction is no more aligned with $u_0$ than the average squared alignment under $G$.

For reference, Jensen's inequality gives the related bound
\[
(u_0^\top m_G)^2
=
\{ \mathbb{E}_G(u_0^\top U)\}^2
\le
\mathbb{E}_G[(u_0^\top U)^2],
\]
where $m_G=\mathbb{E}_G(U)$. This proves non-negativity for the unnormalised mean-summary comparison. For the normalized direction
$u_{\mathrm{pt}}=m_G/\|m_G\|_2$, the comparison above must be imposed or verified separately.
\end{proof}

Assume
\[
\mathbb{E}_G[(u_0^\top U)^2]
>
(u_0^\top u_{\mathrm{pt}})^2 .
\]
Then the gain identity yields
\[
\Delta_{\mathrm{gain}}
=
\|\beta_0\|_2^2
\left[
\mathbb{E}_G[(u_0^\top U)^2]
-
(u_0^\top u_{\mathrm{pt}})^2
\right]
>0.
\]
A sufficient condition for the strict inequality is that $G$ assigns positive mass to directions whose squared alignments with $u_0$ exceed that of the point summary $u_{\mathrm{pt}}$ by enough to make the average squared alignment larger.

\begin{proof}[Proof of the oracle bound]
Let
\[
c_{\mathrm{or}}
=
\sup_{u\in\mathrm{supp}(G)} |u_0^\top u|.
\]
By Theorem~\ref{thm:relative_gain}, the smallest point-wise risk over $\mathrm{supp}(G)$ is achieved by any
\[
u_{\mathrm{or}}
\in
\arg\max_{u\in\mathrm{supp}(G)}
|u_0^\top u|,
\]
and equals
\[
R(u_{\mathrm{or}})
=
\|\beta_0\|_2^2(1-c_{\mathrm{or}}^2).
\]
Moreover,
\[
(u_0^\top U)^2\le c_{\mathrm{or}}^2
\qquad
G\text{-almost surely}.
\]
Therefore
\[
\mathbb{E}_G[(u_0^\top U)^2]\le c_{\mathrm{or}}^2,
\]
and Theorem~\ref{thm:dist_gain} gives
\[
R(G)
=
\|\beta_0\|_2^2
\left\{
1-\mathbb{E}_G[(u_0^\top U)^2]
\right\}
\ge
\|\beta_0\|_2^2(1-c_{\mathrm{or}}^2)
=
R(u_{\mathrm{or}}).
\]
If, in addition,
\[
\mathbb{E}_G[(u_0^\top U)^2]
\ge
(u_0^\top u_{\mathrm{pt}})^2,
\]
then $R(G)\le R(u_{\mathrm{pt}})$. Hence
\[
R(u_{\mathrm{or}})
\le
R(G)
\le
R(u_{\mathrm{pt}}).
\]
\end{proof}

\section{Details of source data generating mechanism in linear and logistic regression example in Section \ref{sec:sim} in the main paper}

\noindent\textbf{Linear regression under unimodal but asymmetric source distribution.}
For each simulation replicate, we first draw the true target direction
$u_0 \sim \mathrm{Unif}(\mathbb{S}^{p-1})$ and set the target coefficient as
$\beta_0=s u_0$. To generate heterogeneous source information, we construct a
discrete directional distribution supported on $M=200$ source directions. The
distribution is designed to be unimodal but asymmetric around the true direction.

Specifically, we first draw a common drift direction $v_0$ uniformly from the
subspace orthogonal to $u_0$, so that $v_0^\top u_0=0$ and $\|v_0\|_2=1$. For
each source draw $m=1,\ldots,M$, we sample a Bernoulli indicator
$Z_m\sim \mathrm{Bernoulli}(0.70)$. If $Z_m=1$, the source direction is sampled
near the truth. We draw an angle
$\theta_m=\min\{|\xi_m|,20^\circ\}$, where
$\xi_m\sim \mathcal{N}(0,8^2)$, draw an independent random unit vector
$v_m\in u_0^\perp$, and set
\[
    U_m=\cos(\theta_m)u_0+\sin(\theta_m)v_m .
\]
The corresponding source scale is drawn as
\[
    R_m\sim \mathrm{LogNormal}(\log 1.2,0.15^2).
\]
If $Z_m=0$, the source direction is drawn from a skewed shoulder along the
common drift direction. We draw
$\theta_m\sim \mathrm{Unif}(18^\circ,32^\circ)$ and set
\[
    U_m=\cos(\theta_m)u_0+\sin(\theta_m)v_0,
    \qquad
    R_m\sim \mathrm{LogNormal}(\log 1.8,0.20^2).
\]
The resulting source coefficient draws are $R_mU_m$, while DUST uses the
corresponding directional empirical distribution
\[
    \mu_s=\frac{1}{M}\sum_{m=1}^M \delta_{U_m}.
\]
This construction produces an informative but asymmetric source distribution:
most source directions are close to the target direction, while a smaller
fraction forms a coherent shoulder along a shared drift direction. Thus, the
source information is useful but not perfectly aligned, making point summaries
slightly biased while allowing distributional directional transfer to adaptively
emphasize the target-supported directions.

\noindent\textbf{Linear regression under bimodal source distribution.}
For each simulation replicate, we first draw the true target direction
$u_0 \sim \mathrm{Unif}(\mathbb{S}^{p-1})$ and set the target coefficient as
$\beta_0=s u_0$ with $s=3$. The target covariates are generated from
$X_i\sim \mathcal{N}(0,\Sigma)$, where $\Sigma_{jk}=0.5^{|j-k|}$, and the response
is generated from the Gaussian linear model
\[
    Y_i=X_i^\top\beta_0+\varepsilon_i,
    \qquad
    \varepsilon_i\sim \mathcal{N}(0,1).
\]

To generate heterogeneous source information, we construct a discrete bimodal
directional distribution supported on $M=200$ source directions. For each source
draw $m=1,\ldots,M$, we sample an indicator
$Z_m\sim \mathrm{Bernoulli}(0.60)$. If $Z_m=1$, the source direction is sampled
from a tight, well-aligned component around the target direction. Specifically,
we draw $\theta_m\sim\mathrm{Unif}(1^\circ,6^\circ)$, sample a random unit vector
$v_m\in u_0^\perp$, and set
\[
    U_m=\cos(\theta_m)u_0+\sin(\theta_m)v_m,
    \qquad
    R_m\sim \mathrm{LogNormal}(\log 1.0,0.08^2).
\]
If $Z_m=0$, the source direction is sampled from a misleading, large-magnitude
component that is nearly orthogonal to the target signal. We draw
$\theta_m\sim\mathrm{Unif}(65^\circ,85^\circ)$, sample
$v_m\in u_0^\perp$, and set
\[
    U_m=\cos(\theta_m)u_0+\sin(\theta_m)v_m,
    \qquad
    R_m\sim \mathrm{LogNormal}(\log 5.0,0.20^2).
\]
The resulting source coefficient draws are $R_mU_m$, while the directional prior
used by DUST is the empirical spherical distribution
\[
    \mu_s=\frac{1}{M}\sum_{m=1}^M\delta_{U_m}.
\]

This construction creates a deliberately challenging source regime: one
component is tightly concentrated near the true target direction, whereas the
other is substantially misaligned and has inflated magnitude. Hence, Euclidean
point summaries can be pulled towards the large-magnitude wrong component, and
single-direction angular summaries can be unstable under the competing source
signals. DUST avoids collapsing the source information to a single point summary
and instead retains the full directional distribution, allowing target-supported
directions to dominate the transfer.

\noindent\textbf{Logistic regression with unimodal asymmetric source distribution.}
For each simulation replicate, we draw the true target direction
$u_0 \sim \mathrm{Unif}(\mathbb{S}^{p-1})$ and set the target coefficient to
$\beta_0=s u_0$ with $s=3$. Covariates are generated as
$X_i\sim \mathcal{N}(0,\Sigma)$, where $\Sigma_{jk}=0.8^{|j-k|}$. The binary
response is generated from the logistic model
\[
    Y_i\mid X_i \sim \mathrm{Bernoulli}(p_i),
    \qquad
    p_i=\{1+\exp(-X_i^\top\beta_0)\}^{-1}.
\]

To generate heterogeneous source information, we construct a discrete, unimodal
but asymmetric directional distribution supported on $M=200$ directions. First,
we draw a common drift direction $v_0$ uniformly from the subspace orthogonal to
$u_0$, so that $v_0^\top u_0=0$ and $\|v_0\|_2=1$. For each source draw
$m=1,\ldots,M$, we sample an indicator $Z_m\sim\mathrm{Bernoulli}(0.60)$. If
$Z_m=1$, the source direction is concentrated near the true target direction:
we draw
\[
    \theta_m=\min\{|\xi_m|,15^\circ\}, \qquad
    \xi_m\sim\mathcal{N}(0,6^2),
\]
sample an independent random unit vector $v_m\in u_0^\perp$, and set
\[
    U_m=\cos(\theta_m)u_0+\sin(\theta_m)v_m,
    \qquad
    R_m\sim\mathrm{LogNormal}(\log 1.1,0.10^2).
\]
If $Z_m=0$, the source direction is drawn from a skewed shoulder along the common
drift direction. We draw
\[
    \theta_m\sim\mathrm{Unif}(22^\circ,42^\circ),
\]
and set
\[
    U_m=\cos(\theta_m)u_0+\sin(\theta_m)v_0,
    \qquad
    R_m\sim\mathrm{LogNormal}(\log 2.2,0.22^2).
\]
The resulting source coefficient draws are $R_mU_m$, while DUST uses the
empirical directional distribution
\[
    \mu_s=\frac{1}{M}\sum_{m=1}^M\delta_{U_m}.
\]

This construction produces source information that is informative but not
perfectly aligned with the target. The concentrated component provides useful
transfer signal, whereas the skewed shoulder introduces systematic directional
bias and scale--direction mismatch. Consequently, Euclidean summaries are
sensitive to inflated magnitudes, and angular point summaries may be pulled
towards the asymmetric shoulder. DUST retains the full distribution over source
directions and allows the target likelihood to emphasise directions that are
predictive for the target task.

\noindent\textbf{Logistic regression with bimodal heterogeneous source distribution.}
For each simulation replicate, we draw the true target direction
$u_0 \sim \mathrm{Unif}(\mathbb{S}^{p-1})$ and set the target coefficient to
$\beta_0=s u_0$ with $s=3$. Covariates are generated as
$X_i\sim \mathcal{N}(0,\Sigma)$, where $\Sigma_{jk}=0.85^{|j-k|}$. The binary
response is generated from the logistic model
\[
    Y_i\mid X_i \sim \mathrm{Bernoulli}(p_i),
    \qquad
    p_i=\{1+\exp(-X_i^\top\beta_0)\}^{-1}.
\]

To generate heterogeneous source information, we construct a discrete bimodal
directional distribution supported on $M=200$ directions. For each source draw
$m=1,\ldots,M$, we sample an indicator $Z_m\sim\mathrm{Bernoulli}(0.60)$. If
$Z_m=1$, the source direction is drawn from a well-aligned component close to
the target direction:
\[
    \theta_m\sim\mathrm{Unif}(1^\circ,6^\circ),
    \qquad
    U_m=\cos(\theta_m)u_0+\sin(\theta_m)v_m,
\]
where $v_m$ is a random unit vector in $u_0^\perp$. The corresponding scale is
\[
    R_m\sim\mathrm{LogNormal}(\log 1.0,0.08^2).
\]
If $Z_m=0$, the source direction is drawn from a misleading component far from
the target direction:
\[
    \theta_m\sim\mathrm{Unif}(65^\circ,85^\circ),
    \qquad
    U_m=\cos(\theta_m)u_0+\sin(\theta_m)v_m,
\]
with inflated magnitude
\[
    R_m\sim\mathrm{LogNormal}(\log 5.0,0.20^2).
\]
The resulting source coefficient draws are $R_mU_m$, and the empirical
directional distribution used by DUST is
\[
    \mu_s=\frac{1}{M}\sum_{m=1}^M\delta_{U_m}.
\]

This source design creates two competing directional components: one tightly
aligned with the target signal and another substantially misaligned with larger
coefficient magnitude. Consequently, Euclidean transfer can be pulled towards
the inflated wrong component, while angular point transfer may be affected by
summarising the two directional components through a single representative
direction. DUST retains the full directional distribution and allows the target
likelihood to favour source directions that are predictive for the target task.

\begin{table}
\tiny
\setlength{\tabcolsep}{3.5pt}

\begin{minipage}{0.49\linewidth}
\centering
\caption{\textbf{Poisson regression.} Test MSE on the linear predictor with standard error in parentheses (lower is better).}
\label{tab:poisson_compact}
\begin{tabular}{cccccc}
\toprule
$n$ & $p$ & Target-only & Euclidean & Angular & DUST \\
\midrule
100 & 10 & 0.075 (0.006) & 0.060 (0.005) & 0.009 (0.001) & \textbf{0.006 (0.001)} \\
100 & 20 & 0.203 (0.013) & 0.151 (0.009) & 0.009 (0.001) & \textbf{0.006 (0.001)} \\
100 & 30 & 0.330 (0.016) & 0.224 (0.009) & 0.009 (0.001) & \textbf{0.005 (0.001)} \\
100 & 40 & 49.774 (48.897) & 0.375 (0.013) & 0.010 (0.002) & \textbf{0.006 (0.001)} \\
100 & 50 & 1.238 (0.082) & 0.534 (0.026) & 0.008 (0.001) & \textbf{0.005 (0.001)} \\
\midrule
200 & 10 & 0.039 (0.003) & 0.035 (0.003) & 0.007 (0.001) & \textbf{0.004 (0.001)} \\
200 & 20 & 0.076 (0.004) & 0.068 (0.004) & 0.007 (0.001) & \textbf{0.004 (0.001)} \\
200 & 30 & 0.130 (0.006) & 0.116 (0.005) & 0.006 (0.001) & \textbf{0.003 (0.001)} \\
200 & 40 & 0.190 (0.009) & 0.163 (0.008) & 0.007 (0.001) & \textbf{0.004 (0.001)} \\
200 & 50 & 0.267 (0.010) & 0.221 (0.008) & 0.007 (0.001) & \textbf{0.003 (0.000)} \\
\midrule
300 & 10 & 0.022 (0.002) & 0.021 (0.001) & 0.005 (0.001) & \textbf{0.002 (0.000)} \\
300 & 20 & 0.049 (0.003) & 0.046 (0.003) & 0.006 (0.001) & \textbf{0.003 (0.000)} \\
300 & 30 & 0.079 (0.004) & 0.073 (0.003) & 0.005 (0.000) & \textbf{0.003 (0.000)} \\
300 & 40 & 0.108 (0.004) & 0.099 (0.003) & 0.006 (0.000) & \textbf{0.003 (0.000)} \\
300 & 50 & 0.144 (0.005) & 0.131 (0.004) & 0.006 (0.000) & \textbf{0.002 (0.000)} \\
\midrule
400 & 10 & 0.019 (0.002) & 0.018 (0.001) & 0.006 (0.001) & \textbf{0.002 (0.000)} \\
400 & 20 & 0.034 (0.002) & 0.033 (0.002) & 0.005 (0.001) & \textbf{0.002 (0.000)} \\
400 & 30 & 0.055 (0.002) & 0.052 (0.002) & 0.005 (0.000) & \textbf{0.002 (0.000)} \\
400 & 40 & 0.079 (0.003) & 0.075 (0.003) & 0.005 (0.000) & \textbf{0.002 (0.000)} \\
400 & 50 & 0.095 (0.003) & 0.089 (0.003) & 0.005 (0.000) & \textbf{0.002 (0.000)} \\
\midrule
500 & 10 & 0.014 (0.001) & 0.013 (0.001) & 0.004 (0.000) & \textbf{0.002 (0.000)} \\
500 & 20 & 0.026 (0.002) & 0.025 (0.001) & 0.004 (0.000) & \textbf{0.002 (0.000)} \\
500 & 30 & 0.040 (0.002) & 0.039 (0.002) & 0.005 (0.000) & \textbf{0.002 (0.000)} \\
500 & 40 & 0.058 (0.003) & 0.056 (0.002) & 0.006 (0.000) & \textbf{0.002 (0.000)} \\
500 & 50 & 0.075 (0.003) & 0.071 (0.003) & 0.005 (0.000) & \textbf{0.002 (0.000)} \\
\bottomrule
\end{tabular}
\end{minipage}
\hfill
\begin{minipage}{0.49\linewidth}
\centering
\caption{\textbf{Quantile regression ($\tau=0.5$).} Test pinball loss with standard error in parentheses (lower is better).}
\label{tab:quantile_compact}
\begin{tabular}{cccccc}
\toprule
$n$ & $p$ & Target-only & Euclidean & Angular & DUST \\
\midrule
100 & 10 & 0.642 (0.003) & 0.627 (0.002) & 0.620 (0.003) & \textbf{0.604 (0.001)} \\
100 & 20 & 0.696 (0.005) & 0.655 (0.003) & 0.622 (0.002) & \textbf{0.604 (0.001)} \\
100 & 30 & 0.775 (0.008) & 0.698 (0.005) & 0.620 (0.002) & \textbf{0.602 (0.001)} \\
100 & 40 & 0.869 (0.011) & 0.753 (0.006) & 0.625 (0.002) & \textbf{0.605 (0.001)} \\
100 & 50 & 0.963 (0.017) & 0.809 (0.010) & 0.625 (0.002) & \textbf{0.605 (0.002)} \\
\midrule
200 & 10 & 0.618 (0.002) & 0.612 (0.002) & 0.618 (0.002) & \textbf{0.603 (0.001)} \\
200 & 20 & 0.643 (0.002) & 0.631 (0.002) & 0.620 (0.002) & \textbf{0.603 (0.001)} \\
200 & 30 & 0.677 (0.004) & 0.647 (0.002) & 0.618 (0.002) & \textbf{0.600 (0.001)} \\
200 & 40 & 0.700 (0.004) & 0.657 (0.002) & 0.621 (0.002) & \textbf{0.605 (0.001)} \\
200 & 50 & 0.730 (0.004) & 0.673 (0.003) & 0.620 (0.002) & \textbf{0.602 (0.001)} \\
\midrule
300 & 10 & 0.613 (0.001) & 0.614 (0.001) & 0.613 (0.002) & \textbf{0.601 (0.001)} \\
300 & 20 & 0.628 (0.002) & 0.621 (0.002) & 0.616 (0.002) & \textbf{0.603 (0.001)} \\
300 & 30 & 0.638 (0.002) & 0.626 (0.002) & 0.616 (0.002) & \textbf{0.602 (0.001)} \\
300 & 40 & 0.664 (0.003) & 0.638 (0.002) & 0.623 (0.002) & \textbf{0.603 (0.001)} \\
300 & 50 & 0.679 (0.003) & 0.647 (0.002) & 0.621 (0.002) & \textbf{0.604 (0.001)} \\
\midrule
400 & 10 & 0.607 (0.001) & 0.612 (0.001) & 0.620 (0.003) & \textbf{0.602 (0.001)} \\
400 & 20 & 0.622 (0.002) & 0.618 (0.001) & 0.620 (0.002) & \textbf{0.601 (0.001)} \\
400 & 30 & 0.630 (0.001) & 0.621 (0.001) & 0.616 (0.002) & \textbf{0.602 (0.001)} \\
400 & 40 & 0.643 (0.002) & 0.626 (0.002) & 0.619 (0.002) & \textbf{0.602 (0.001)} \\
400 & 50 & 0.655 (0.002) & 0.634 (0.001) & 0.621 (0.002) & \textbf{0.601 (0.001)} \\
\midrule
500 & 10 & 0.611 (0.001) & 0.611 (0.002) & 0.615 (0.002) & \textbf{0.600 (0.001)} \\
500 & 20 & 0.615 (0.001) & 0.614 (0.001) & 0.617 (0.001) & \textbf{0.601 (0.001)} \\
500 & 30 & 0.623 (0.001) & 0.617 (0.001) & 0.618 (0.002) & \textbf{0.602 (0.001)} \\
500 & 40 & 0.635 (0.001) & 0.624 (0.001) & 0.620 (0.002) & \textbf{0.602 (0.001)} \\
500 & 50 & 0.643 (0.002) & 0.627 (0.002) & 0.620 (0.002) & \textbf{0.603 (0.001)} \\
\bottomrule
\end{tabular}
\end{minipage}

\end{table}

\section{Poisson and quantile regression: directional transfer under heterogeneous source structure}
\noindent\textbf{Poisson regression: directional transfer under heterogeneous source structure.}
We consider a count regression setting generated from the log-linear model
$Y_i \mid X_i \sim \mathrm{Poisson}(\lambda_i)$ with
$\log \lambda_i = X_i^\top \beta_0$. The dimension is
$p \in \{10,20,30,40,50\}$ and the sample size is
$n \in \{100,200,300,400,500\}$. Covariates are generated as
$X_i \sim \mathcal{N}(0,\Sigma)$ with
$\Sigma_{jk}=0.85^{|j-k|}$, inducing moderate correlation among features. The
true parameter is $\beta_0 = s u_0$, where
$u_0 \sim \mathrm{Unif}(\mathbb{S}^{p-1})$ and $s=1$, so that the signal remains
moderate under the exponential link. Source information is generated from a
heterogeneous directional distribution containing well-aligned directions
close to $u_0$ and moderately misaligned directions with inflated magnitudes.
This induces scale--direction confounding: Euclidean transfer is affected by
magnitude inflation, while angular transfer reduces scale sensitivity but
collapses the source distribution to a single direction. DUST instead uses the
full directional distribution. Performance is measured using test MSE on the linear
predictor,
$\mathbb{E}\{(X_{\mathrm{new}}^\top\hat\beta -
X_{\mathrm{new}}^\top\beta_0)^2\}$, averaged over Monte Carlo replications.

\noindent\textbf{Quantile regression: directional transfer under heterogeneous source structure.}
We next consider median regression under heavy-tailed noise,
$Y_i = X_i^\top\beta_0+\varepsilon_i$, where $\varepsilon_i$ is generated from a
difference of exponentials distribution. The same $(n,p)$ grid and covariance
structure are used as in the Poisson experiment. Performance is evaluated using
the pinball loss at $\tau=0.5$. The source distribution again contains a
mixture of well-aligned and misaligned directions with scale heterogeneity.
This setting is less sensitive to extreme responses than squared-error
regression, but remains challenging because point-based transfer methods must
summarize heterogeneous source information by a single coefficient vector or a
single direction. DUST  aggregates over plausible
directions and allows the target loss to determine which source directions are
most predictive.

The Poisson results show a pronounced benefit from directional distributional transfer. DUST uniformly achieves the smallest test MSE on the log-mean scale. The gap relative to Euclidean transfer is large because Euclidean shrinkage is sensitive to source magnitude inflation, while angular transfer is substantially more stable but still loses information by selecting a single source direction. The target-only estimator can become unstable in small-sample, higher-dimensional regimes, as seen for $n=100,p=40$, whereas DUST remains stable by borrowing from directions that are predictive under the target likelihood.

The quantile regression results show the same qualitative ordering but with smaller absolute gains. This is expected because the median pinball loss is already robust to heavy-tailed noise, reducing the scope for improvement. Nevertheless, DUST is consistently best across all $(n,p)$ settings. The improvements over angular transfer are modest but systematic, supporting the central claim that distributional transfer is useful even when the baseline loss function is robust.

\begin{algorithm}[t]
\footnotesize
\caption{Quantile regression. Posterior sampler for  directional transfer learning with angular cone prior}
\label{alg:gibbs_cone_quantile}
\KwIn{Target data $(Y,X)$; quantile level $\tau\in(0,1)$; source directional prior $\mu_s$; hyperparameters $(c,\varepsilon,\tau_a^2,\sigma_0^2)$}
\KwOut{Posterior samples $\{\beta^{(m)},U^{(m)},a^{(m)},\theta^{(m)}\}_{m=1}^M$}

\BlankLine
\textbf{Working likelihood:}
Use the asymmetric Laplace working likelihood
\[
p_\tau(Y_i\mid X_i,\beta,\sigma_0)
\propto
\exp\left\{
-\frac{1}{\sigma_0}\rho_\tau(Y_i-X_i^\top\beta)
\right\},
\qquad
\rho_\tau(r)=r\{\tau-1(r<0)\}.
\]

\BlankLine
\textbf{Initialization:}
Initialize $U^{(0)}\in S^{p-1}$, $a^{(0)}\in\mathbb{R}$, and $\theta^{(0)}\in\mathbb{R}^{p-1}$.

\For{$m=1,\dots,M$}{
\BlankLine
\textbf{1. Orthogonal reparameterization.}  
Given $U^{(m-1)}$, construct $Q_{U^{(m-1)}}$ with columns spanning $\mathrm{span}(U^{(m-1)})^\perp$.  
Set $x_U = XU^{(m-1)}$ and $Z_U = XQ_{U^{(m-1)}}$.

\BlankLine
\textbf{2. Update $\theta$.}  
Sample $\theta^{(m)}$ from the conditional density
\[
p(\theta\mid a^{(m-1)},U^{(m-1)},Y)
\propto
\exp\left[
-\frac{1}{\sigma_0}
\sum_{i=1}^n
\rho_\tau\!\left(Y_i-a^{(m-1)}x_{U,i}-Z_{U,i}^\top\theta\right)
-\frac{a^{(m-1)2}+\varepsilon}{2c^2}\|\theta\|_2^2
\right].
\]
Update this log-concave but non-Gaussian block using elliptical slice sampling, random-walk Metropolis, or a proximal/Laplace proposal.

\BlankLine
\textbf{3. Update $a$.}  
Sample $a^{(m)}$ from the one-dimensional density
\[
p(a\mid\theta^{(m)},U^{(m-1)},Y)
\propto
\exp\{\ell_\tau(a)\},
\]
where
\[
\ell_\tau(a)
=
-\frac{1}{\sigma_0}
\sum_{i=1}^n
\rho_\tau\!\left(Y_i-a x_{U,i}-Z_{U,i}^\top\theta^{(m)}\right)
-\frac{a^2}{2\tau_a^2}
+\frac{p-1}{2}\log(a^2+\varepsilon)
-\frac{a^2+\varepsilon}{2c^2}\|\theta^{(m)}\|_2^2.
\]
Update using slice sampling or adaptive rejection sampling.

\BlankLine
\textbf{4. Update $U$.}

\emph{If $\mu_s$ is discrete} on $\{U^{(1)},\dots,U^{(M_s)}\}$ with weights $\{\omega_j\}$, sample
\[
\mathbb{P}\big(U^{(m)}=U^{(j)}\mid\cdot\big)
\propto
\omega_j
\exp\left[
-\frac{1}{\sigma_0}
\sum_{i=1}^n
\rho_\tau\!\left(
Y_i-a^{(m)}x_{U^{(j)},i}
-Z_{U^{(j)},i}^\top\theta^{(m)}
\right)
-\frac{a^{(m)2}+\varepsilon}{2c^2}\|\theta^{(m)}\|_2^2
\right].
\]

\emph{If $\mu_s$ is continuous}, propose $U'$ on $S^{p-1}$ and accept with Metropolis probability
\[
\alpha(U\to U')
=
1\wedge
\frac{
p(U')\,
\exp\!\left[-\sigma_0^{-1}\sum_{i=1}^n
\rho_\tau\{Y_i-a^{(m)}x_{U',i}-Z_{U',i}^\top\theta^{(m)}\}\right]
}{
p(U^{(m-1)})\,
\exp\!\left[-\sigma_0^{-1}\sum_{i=1}^n
\rho_\tau\{Y_i-a^{(m)}x_{U^{(m-1)},i}-Z_{U^{(m-1)},i}^\top\theta^{(m)}\}\right]
}
\frac{q(U^{(m-1)}\mid U')}{q(U'\mid U^{(m-1)})}.
\]

\BlankLine
\textbf{5. Recover $\beta$.}  
Set
\[
\beta^{(m)} = a^{(m)}U^{(m)} + Q_{U^{(m)}}\theta^{(m)}.
\]
}

\BlankLine
\Return Posterior samples $\{\beta^{(m)},U^{(m)},a^{(m)},\theta^{(m)}\}_{m=1}^M$.
\end{algorithm}

\clearpage
\section*{NeurIPS Paper Checklist}

\begin{enumerate}

\item {\bf Claims}
    \item[] Question: Do the main claims made in the abstract and introduction accurately reflect the paper's contributions and scope?
    \item[] Answer:  \answerYes{}.
    \item[] Justification: Main claims made in the abstract and introduction accurately reflect the paper's contributions and scope.

\item {\bf Limitations}
    \item[] Question: Does the paper discuss the limitations of the work performed by the authors?
    \item[] Answer:  \answerYes{}.
    \item[] Justification: We theoretically characterize regimes under which the proposed framework yields gains relative to these baselines. 

\item {\bf Theory assumptions and proofs}
    \item[] Question: For each theoretical result, does the paper provide the full set of assumptions and a complete (and correct) proof?
    \item[] Answer: \answerYes{}.
    \item[] Justification:  For each theoretical result, the paper provides the full set of assumptions and a complete (and correct) proof.

    \item {\bf Experimental result reproducibility}
    \item[] Question: Does the paper fully disclose all the information needed to reproduce the main experimental results of the paper to the extent that it affects the main claims and/or conclusions of the paper (regardless of whether the code and data are provided or not)?
    \item[] Answer: \answerYes{}.
    \item[] Justification: The paper fully disclose all the information needed to reproduce the main experimental results of the paper to the extent that it affects the main claims and/or conclusions of the paper.

\item {\bf Open access to data and code}
    \item[] Question: Does the paper provide open access to the data and code, with sufficient instructions to faithfully reproduce the main experimental results, as described in supplemental material?
    \item[] Answer:  \answerYes{}.
    \item[] Justification: The paper provides open access to the data and code, with sufficient instructions to faithfully reproduce the main experimental results.

\item {\bf Experimental setting/details}
    \item[] Question: Does the paper specify all the training and test details (e.g., data splits, hyperparameters, how they were chosen, type of optimizer) necessary to understand the results?
    \item[] Answer: \answerYes{}.
    \item[] Justification: The paper specifies all the training and test details

\item {\bf Experiment statistical significance}
    \item[] Question: Does the paper report error bars suitably and correctly defined or other appropriate information about the statistical significance of the experiments?
    \item[] Answer:  \answerYes{}.
    \item[] Justification: The paper reports error bars for the experiments.

\item {\bf Experiments compute resources}
    \item[] Question: For each experiment, does the paper provide sufficient information on the computer resources (type of compute workers, memory, time of execution) needed to reproduce the experiments?
    \item[] Answer:  \answerYes{}.
    \item[] Justification: All experiments were conducted on a standard personal laptop ( a Dell XPS 15 equipped with an Intel Core i7 processor with 8 cores and 32 GB RAM), without the use of GPU acceleration. The paper provides sufficient implementation details, and the computational requirements are modest, with runtimes that are readily reproducible on comparable consumer-grade hardware.
    
\item {\bf Code of ethics}
    \item[] Question: Does the research conducted in the paper conform, in every respect, with the NeurIPS Code of Ethics \url{https://neurips.cc/public/EthicsGuidelines}?
    \item[] Answer: \answerYes{}.
    \item[] Justification: The research conducted in the paper conform, in every respect, with the NeurIPS Code of Ethics.

\item {\bf Broader impacts}
    \item[] Question: Does the paper discuss both potential positive societal impacts and negative societal impacts of the work performed?
    \item[] Answer: \answerNA{}.
    \item[] Justification: There is no societal impact of the work performed.
    
\item {\bf Safeguards}
    \item[] Question: Does the paper describe safeguards that have been put in place for responsible release of data or models that have a high risk for misuse (e.g., pre-trained language models, image generators, or scraped datasets)?
    \item[] Answer:  \answerNA{}.
    \item[] Justification: The proposed model do not have a high risk for misuse.

\item {\bf Licenses for existing assets}
    \item[] Question: Are the creators or original owners of assets (e.g., code, data, models), used in the paper, properly credited and are the license and terms of use explicitly mentioned and properly respected?
    \item[] Answer:  \answerNA{}.
    \item[] Justification: The authors are the owners of the assets, with the exception of the datasets, which are publicly available and open-source.

\item {\bf New assets}
    \item[] Question: Are new assets introduced in the paper well documented and is the documentation provided alongside the assets?
    \item[] Answer: \answerYes{}.
    \item[] Justification: The details of the model are described in the paper. The codes are provided as supplementary material.

\item {\bf Crowdsourcing and research with human subjects}
    \item[] Question: For crowdsourcing experiments and research with human subjects, does the paper include the full text of instructions given to participants and screenshots, if applicable, as well as details about compensation (if any)? 
    \item[] Answer:  \answerNA{}.
    \item[] Justification: The paper does not involve crowdsourcing nor research with human subjects. Open-source datasets are used for demonstrations.
    \item[] Guidelines: The paper does not involve crowdsourcing experiments and research with human subjects.
    \begin{itemize}
        \item The answer \answerNA{} means that the paper does not involve crowdsourcing nor research with human subjects. 
        \item Including this information in the supplemental material is fine, but if the main contribution of the paper involves human subjects, then as much detail as possible should be included in the main paper. 
        \item According to the NeurIPS Code of Ethics, workers involved in data collection, curation, or other labor should be paid at least the minimum wage in the country of the data collector. 
    \end{itemize}

\item {\bf Institutional review board (IRB) approvals or equivalent for research with human subjects}
    \item[] Question: Does the paper describe potential risks incurred by study participants, whether such risks were disclosed to the subjects, and whether Institutional Review Board (IRB) approvals (or an equivalent approval/review based on the requirements of your country or institution) were obtained?
    \item[] Answer: \answerNA{}.
    \item[] Justification: The paper does not involve crowdsourcing nor research with human subjects.
    \item[] Guidelines: Not applicable. 
    \begin{itemize}
        \item The answer \answerNA{} means that the paper does not involve crowdsourcing nor research with human subjects.
        \item Depending on the country in which research is conducted, IRB approval (or equivalent) may be required for any human subjects research. If you obtained IRB approval, you should clearly state this in the paper. 
        \item We recognize that the procedures for this may vary significantly between institutions and locations, and we expect authors to adhere to the NeurIPS Code of Ethics and the guidelines for their institution. 
        \item For initial submissions, do not include any information that would break anonymity (if applicable), such as the institution conducting the review.
    \end{itemize}

\item {\bf Declaration of LLM usage}
    \item[] Question: Does the paper describe the usage of LLMs if it is an important, original, or non-standard component of the core methods in this research? Note that if the LLM is used only for writing, editing, or formatting purposes and does \emph{not} impact the core methodology, scientific rigor, or originality of the research, declaration is not required.
    \item[] Answer: \answerNA{}.
    \item[] Justification: The core method development in this research does not involve LLMs as any important, original, or non-standard components.

\end{enumerate}
\end{document}